\documentclass{bmvc2k}

\newif\ifsubmission
\submissiontrue

\usepackage{listings}
\usepackage{amsfonts}
\usepackage{amsmath}
\usepackage{multirow}
\usepackage{soul}  
\usepackage{graphicx}

\allowdisplaybreaks

\newcommand{\mv}{\mathbf}

\usepackage{array}
\newcolumntype{P}[1]{>{\centering\arraybackslash}p{#1}}

\title{Why is the Mahalanobis Distance Effective for Anomaly Detection?}

\addauthor{Ryo Kamoi}{}{2}
\addauthor{Kei Kobayashi}{}{1}

\addinstitution{
 Department of Mathematics\\
 Keio University
}
\addinstitution{
 Department of Computer Science\\
 The University of Tokyo
}

\runninghead{Kamoi \& Kobayashi}{Why is Mahalanobis Effective for Anom Detection?}

\begin{document}
\maketitle

\begin{abstract}
The Mahalanobis distance-based confidence score,
a recently proposed anomaly detection method for pre-trained neural classifiers,
achieves state-of-the-art performance
on both out-of-distribution (OoD) and adversarial examples detection.
This work analyzes why this method exhibits such strong performance in practical settings
while imposing an implausible assumption; namely,
that class conditional distributions of pre-trained features have tied covariance.
Although the Mahalanobis distance-based method is claimed to be motivated by
classification prediction confidence,
we find that its superior performance stems from
information not useful for classification.
This suggests that the reason the Mahalanobis confidence score works so well is mistaken,
and makes use of different information from ODIN,
another popular OoD detection method based on prediction confidence.
This perspective motivates us to combine these two methods,
and the combined detector exhibits improved performance and robustness.
These findings provide insight into the behavior of
neural classifiers in response to anomalous inputs.
\end{abstract}

\section{Introduction}
Modern neural networks often exhibit unexpected behavior
to inputs dissimilar from training data
\citep{Gal2016thesis,Hendrycks2017a}.
It is also known that neural classifiers are easily fooled by 
small adversarial perturbations to inputs
\citep{Szegedy2013IntriguingNetworks,Goodfellow2015ExplainingAdversarial,Nguyen2015DeepImages}.
Since these problems pose a serious threat to the safety of machine learning systems,
anomaly detection methods such as 
out-of-distribution (OoD) and adversarial examples detection
have attracted considerable attention
\citep{Hendrycks2017a,Liang2018EnhancingNetworks,Pimentel2014ADetection,Li2017AdversarialStatistics}.
However, the behavior of neural networks when fed anomalous inputs is not well understood.

Distance-based methods are a primitive yet popular approach to anomaly detection.
One way to apply this approach to neural networks
is to use lower dimensional representations for data.
Previous studies on OoD
\citep{Song2017AData, Guo2018AnNeighbor}
and adversarial examples detection
\citep{Feinman2017DetectingArtifacts, Ma2018CharacterizingDimensionality, Gilmer2018TheExamples, Papernot2018DeepLearning}
have used the Euclidean distance of intermediate representations in neural networks.
In this line of research,
recent work on anomaly detection for
pre-trained neural classifiers \citep{Lee2018AAttacks}
has shown that the Mahalanobis distance-based confidence score
outperforms methods based on the Euclidean distance.
This method achieves state-of-the-art performance
in OoD and adversarial examples detection,
and is regarded as the new standard in anomaly detection on pre-trained neural classifiers
\citep{Ren2019LikelihoodDetection, Rajendran2019AccurateEstimation, Tardy2019UncertaintyMammograms, Roady2019AreDatasets, Wang2019InterpretableSubnetwork, Macedo2019DistinctionLoss, Ahuja2019ProbabilisticDetection, Cohen2019DetectingNeighbors}.
However,
why it works so well has not been sufficiently scrutinized.

The Mahalanobis confidence score \citep{Lee2018AAttacks} assumes that
the intermediate features of pre-trained neural classifiers
follow class conditional Gaussian distributions
whose covariances are tied for all distributions,
and the confidence score for a new input is defined as
the Mahalanobis distance from the closest class conditional distribution.
Our work is motivated by the question of
why the Mahalanobis confidence score exhibits strong performance 
while imposing this unreasonable and restrictive assumption of tied covariance.

We reveal that the reasons why the Mahalanobis confidence score
can detect anomalous samples effectively has been misunderstood.
This confidence score is derived from prediction confidence for classification,
a similar idea to ODIN \citep{Hendrycks2017a} which uses
the output of the softmax function as a confidence score.
However, our analysis shows that
the information in intermediate features that is not utilized as much for classification
is the main contributor to state-of-the-art performance of the method.
We propose a simpler method named the marginal Mahalanobis confidence score.
This method approximates the penultimate feature distribution with a single Gaussian distribution,
and uses the Mahalanobis distance from the mean as confidence scores.
In most settings of anomaly detection,
our method which does not use class information at all,
is competitive with the original confidence score.

\begin{figure}[tb]
\vskip 0.2in
\center
\begin{tabular}{P{.45\linewidth}P{.45\linewidth}}
\bmvaHangBox{\includegraphics[height=1in]{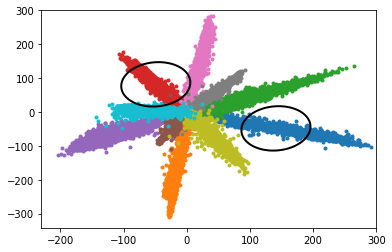}} &
\bmvaHangBox{\includegraphics[height=1in]{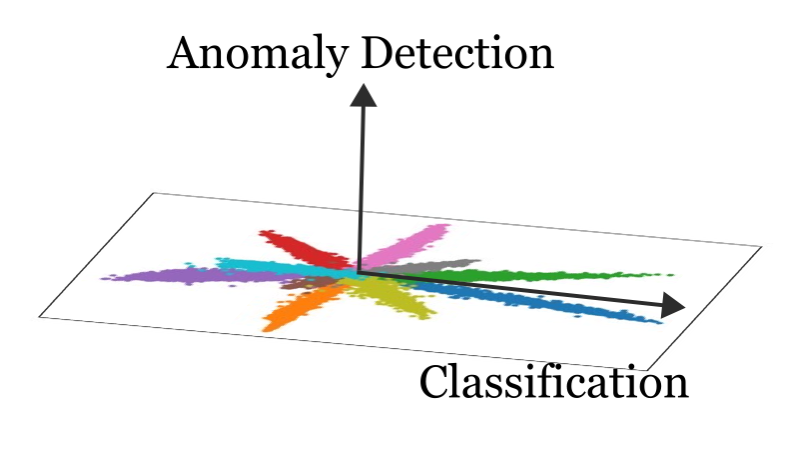}} \\
(a) & (b) \\
\end{tabular}
\vskip 1em
\caption{
Motivation and proposition of this work.
(a) Final features of a convolutional neural network classifier
trained on MNIST.
The final feature space is set to two dimensions.
Black ovals are contours of the Mahalanobis distance
under the tied covariance assumption.
The Mahalanobis distance does not properly describe
the class conditional distributions in this example.
However, this score by \citet{Lee2018AAttacks}
achieves state-of-the-art performance on
anomaly detection under practical settings.
(b) Our analysis suggests that directions with small explained variance,
while containing relatively less
information relevant to classification,
strongly contribute to anomaly detection in high-dimensional cases.
\label{fig:summary}}
\end{figure}
\begin{table}[tb]
\center
\footnotesize
\begin{tabular}{|c|c|c|c|}
\hline
Feature & classification confidence & others & combined \\
\hline \hline
\multirow{2}{*}{OoD detection} & Baseline \citep{Hendrycks2017a}, ODIN \citep{Liang2018EnhancingNetworks} &
\multirow{2}{*}{Mahalanobis (\citep{Lee2018AAttacks}, ours)} & Mahalanobis \\
& \st{Mahalanobis} \citep{Lee2018AAttacks} &  & + ODIN (ours) \\ \hline
Adversarial & Bayesian Uncertainty \citep {Feinman2017DetectingArtifacts} 
& Mahalanobis (\citep{Lee2018AAttacks}, ours), LID \citep{Ma2018CharacterizingDimensionality} &
Uncertainty \\
examples detection & \st{Mahalanobis} \citep{Lee2018AAttacks} & 
\citet{Li2017AdversarialStatistics}, Kernel Density \citep{Feinman2017DetectingArtifacts} & 
+ Density \cite{Feinman2017DetectingArtifacts} \\ \hline
\end{tabular}
\vskip 1em
\caption{Categorization of OoD and adversarial examples detection methods 
for pre-trained neural classifiers based on our analysis.
The Mahalanobis confidence score \citep{Lee2018AAttacks} had been believed
to use classification prediction confidence,
but our analysis shows otherwise.
Motivated by this categorization, we propose a combined method for OoD detection.}
\label{table:classify_detection_methods}
\end{table}

In addition,
as a consequence of our analysis,
the categorization in Table~\ref{table:classify_detection_methods} suggests that
the Mahalanobis confidence score and ODIN, two popular OoD detection methods,
distinguish between in-distribution and OoD data in different ways.
Thus, we propose combining the Mahalanobis score and ODIN for OoD detection.
We experimentally show this idea improves OoD detection performance and robustness.

\vskip .5em
\noindent
{\bf Contribution}
The Mahalanobis confidence score \citep{Lee2018AAttacks} had been believed
to function based on the prediction confidence of classifiers.
However, we reveal that the Mahalanobis confidence score is 
a powerful variant of a method detecting adversarial examples
by using properties observed by \citet{Li2017AdversarialStatistics},
unrelated to prediction confidence.
Our analysis deepens our understanding of the Mahalanobis confidence score,
the new standard for anomaly detection,
and draws connections with previous studies on OoD and adversarial examples detection.
Finally, based on this new perspective,
we propose combining the Mahalanobis confidence score and ODIN.

\section{Related Work}
{\bf High-Dimensional OoD Detection}
\citet{Hendrycks2017a} proposed a baseline method for pre-trained neural classifiers
using the output of the softmax function as a confidence score,
and \citet{Liang2018EnhancingNetworks} significantly improved upon this.
\citet{Hendrycks2019DeepExposure} proposed to use OoD data during training.
An approach related to this paper is the use of
lower dimensional representations in neural networks.
\citet{Mandelbaum2017Distance-basedClassifiers} proposed a confidence score based on
local density estimation in the representation space.
There is some work using intermediate features of autoencoders as well
\citep{Sabokrou2018AdversariallyDetection, Denouden2018ImprovingDistance, Perera2019OCGAN:Representations}.

\noindent
{\bf Adversarial Examples Detection}
Adversarial examples
\citep{Szegedy2013IntriguingNetworks,Goodfellow2015ExplainingAdversarial}
are a major concern of neural networks.
A popular approach to avoid this problem is
to make neural networks robust against these attacks with methods such as
adversarial training \citep{Goodfellow2015ExplainingAdversarial}.
Another approach to mitigate adversarial attacks is adversarial examples detection.
As methods using statistical properties in feature space,
\citet{Li2017AdversarialStatistics} proposed
features acquired via principal component analysis (PCA) on intermediate representations,
\citet{Ma2018CharacterizingDimensionality} proposed usage of
the Local Intrinsic Dimensionality,
and \citet{Feinman2017DetectingArtifacts} utilized kernel density estimation.
\citet{Feinman2017DetectingArtifacts} also proposed using
Bayesian uncertainty, a prediction confidence.
However,
\citet{Carlini2017AdversarialMethods} have reported that
these detection methods can be bypassed relatively easily.

\section{Background}
We describe the Mahalanobis distance-based confidence score.
\citet{Lee2018AAttacks} proposed a confidence score for anomaly detection
on pre-trained neural classifiers
based on the class-conditional Mahalanobis distance with the assumption of tied covariance
in representation space.
This method is motivated by an induced generative classifier.
Here, we consider a generative classifier
with the assumption that the class priors follow categorical distributions
$P(t=c) = \beta_c / \sum_{c'}\beta_{c'}$ and
class-conditional distributions follow Gaussian distributions
with tied covariance $\mathcal N(\mathbf x|\mu_c, \Sigma)$.
Then, its posterior distribution $P(t=c|\mathbf x)$ can be
represented in the following manner:
\begin{equation}
P(t=c|\mathbf x) = \frac{\exp (
\mu_c^T \Sigma^{-1}\mathbf x - \frac{1}{2}\mu_c^T \Sigma^{-1}\mu_c + \log \beta_c)}
{\sum_{c'} \exp (\mu_{c'}^T\Sigma^{-1}\mathbf x
 - \frac{1}{2}\mu_{c'}^T \Sigma^{-1}\mu_{c'} + \log \beta_{c'})}.
\end{equation}
This posterior distribution can be expressed with a softmax classifier by setting
$\mu_c^T \Sigma^{-1}$ and 
$-\frac{1}{2}\mu_c^T \Sigma^{-1}\mu_c + \log \beta_c$ 
to be the weights and biases.
This observation leads to the idea that
the final features of pre-trained neural classifier might
follow a class-conditional Gaussian distribution with tied covariance,
and the Mahalanobis distances from class means can be used as confidence scores.
Given training data $\{(\mv x_1, t_1), \ldots, (\mv x_n, t_n) \}$,
the parameters of the generative classifier can be estimated as follows:
\begin{equation}
    \hat \mu_c = \frac{1}{n_c} \sum_{i:t_i=c} f(\mv x_i), \hspace{1em}
    \hat \Sigma = \frac{1}{n} \sum_c \sum_{i:t_i=c} (f(\mv x_i) - \hat \mu_c)(f(\mv x_i) - \hat \mu_c)^T
\end{equation}
where $f(\cdot)$ denotes the output of the penultimate layer,
and $n_c$ is the number of training examples with label $c$.
The Mahalanobis distance-based score is calculated as 
the minimum squared Mahalanobis distance from the class means:
\begin{equation}
    M(\mv x) = \max_c -(f(\mv x) - \hat \mu_c)^T \hat \Sigma^{-1} (f(\mv x) - \hat \mu_c).
\end{equation}

To improve detection performance,
the authors proposed
input pre-processing similar to \citet{Liang2018EnhancingNetworks} and
a feature ensemble consisting of
logistic regression using confidence scores calculated on all layers.

\section{Analysis and Proposition}
{\bf The Unreasonable Assumption Motivating Our Analysis}
Our analysis is motivated by an unreasonable assumption;
\citet{Lee2018AAttacks} assumes that
class conditional distributions of the final features have tied covariance.
This assumption is not consistent with well-known properties of
representation spaces of neural classifiers.
The standard framework of neural classifiers
with a softmax function following a fully connected layer
evaluates similarity based on inner products.
Thus,
the final features are radially distributed 
as the decision rule is largely determined by angular similarity
\citep{Wen2016ARecognition, Liu2016Large-MarginNetworks}
as shown in Figure~\ref{fig:summary} (a),
so they would not have tied covariance.
\citet{Ahuja2019ProbabilisticDetection} has already pointed out this problem.
However, to the best of our knowledge,
this is the first work analyzing why
the Mahalanobis confidence score detects anomalous samples effectively
despite imposing this unrealistic assumption.

\noindent
{\bf Hypothesis}
Since Figure~\ref{fig:summary} (a) suggests that the Mahalanobis confidence score
is not effective for low-dimensional features,
we hypothesize that it utilizes a property peculiar to high-dimensional feature spaces.
In high-dimensional representation spaces,
a small number of directions will explain almost all the variance in training data,
and the other directions may not contribute to classification.
In the subspace containing useful information for classification,
the in-distribution features are expected to be radially distributed as explained above,
which would not be well fitted by class conditional Gaussian distributions with tied covariance.
Therefore, in high-dimensional cases,
we hypothesize that the directions which are not important for
the classification task provide useful information for anomaly detection.

In this section, we verify this hypothesis by
proposing two methods called partial and marginal Mahalanobis distance.
Finally, we propose a simple method to improve detection performance
based on our analysis.

\subsection{Observation on Intermediate Features}
\label{section:observation_on_intermediate_features}
\begin{figure}[t]
\center
\begin{tabular}{P{.45\linewidth}P{.45\linewidth}}
\bmvaHangBox{\includegraphics[height=1.in]{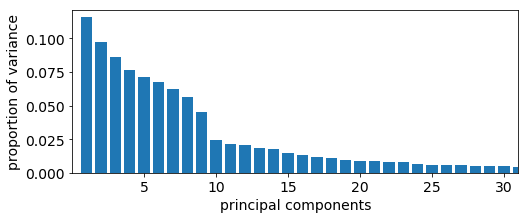}} &
\bmvaHangBox{\includegraphics[height=1.in]{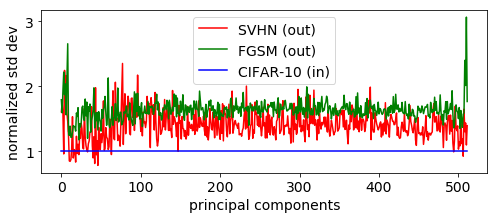}} \\
(a) CIFAR-10 (training data) &(b) Normalized Std. Deviation \\
\end{tabular}
\vskip 1em
\caption{PCA on the final features of ResNet trained on CIFAR-10.
The nearest class mean is subtracted from each input.
(a) The proportion of variance of training data explained by each principal component.
The first several principal components explain almost all of the variance.
(b) Standard deviation of SVHN (OoD inputs) and FGSM
\citep{Goodfellow2015ExplainingAdversarial} (adversarial examples)
normalized by that of in-distribution data.
A similar observation was provided by \citet{Li2017AdversarialStatistics} for adversarial examples.
We argue the Mahalanobis confidence score utilizes this property to
distinguish between in-distribution and anomalous inputs.
\label{fig:partial_pca}}
\end{figure}
{\bf Motivating Observation}
Figure~\ref{fig:partial_pca} (a) shows the proportion of variance when
PCA is applied to the final features of a ResNet classifier trained on CIFAR-10.
Here, we subtract the nearest class mean in terms of the Mahalanobis distance from each input,
which corresponds to \citet{Lee2018AAttacks}'s method.
A small number of principal components explain almost all variance in the data,
so most of the directions may not contain useful information for classification.
However,
Figure~\ref{fig:partial_pca} (b) shows that OoD data and adversarial examples have larger variance
than in-distribution data even on principal components with small explained variance,
which may not be expected to contain useful information for classification.
In other words, the subspace spanned by principal components with small explained variance also
contains information useful for detecting anomalous inputs from in-distribution data.
\citet{Li2017AdversarialStatistics} reported a similar observation for adversarial examples,
so this analysis shows that this phenomenon also occurs on OoD data.

\noindent
{\bf Classification vs. Detection}
We conduct a simple experiment to verify that
information in the subspace spanned by principal components with small explained variance
does not contribute to classification.
We apply PCA to the final features,
and evaluate classification accuracies using subsets of the principal components.
Logistic regression on the final features of neural classifiers trained on CIFAR-10
achieves $94.1\%$ accuracy when all principal components are used.
A classifier using the first $9$ principal components 
with large proportion of variance achieves $94.1\%$,
but one using the $10$-th to $512$-th components only achieves $20.5\%$.
These results suggest that principal components with small explained variance
do not contain information critical to classification,
while Figure~\ref{fig:partial_pca} suggest they provide useful information for anomaly detection.
Note that we do not subtract class means in this experiment here,
so this result corresponds to the marginal Mahalanobis distance explained later.

\subsection{Partial Mahalanobis Distance\label{subsection:partial}}
{\bf Background}
The Mahalanobis distance can be alternatively represented with
the eigenvectors $\mathbf u_i$ of the covariance matrix $\Sigma$ as
$\Delta^2 = (\mathbf x-\mu)^T \Sigma^{-1} (\mathbf x- \mu)
= \sum_{i=1}^N y_i^2 / \lambda_i$,
where $N$ is the dimension of data,
$\lambda_i$ is the $i$-th eigenvalue,
and $y_i = \mathbf u_i^T (\mathbf x-\mu)$ \citep{Murphy2012MachinePerspective}.
Note that the eigenvectors and eigenvalues of data sets can be interpreted as
the principal components and explained variance of PCA,
so this formulation shows the relationship between
the Mahalanobis distance and PCA.

\noindent
{\bf Definition}
To verify our hypothesis,
we want to evaluate detection performance of the Mahalanobis score
only using information that is not critical for classification task.
Motivated by the fact that the squared Mahalanobis distance is
equivalent to the squared sum of principal component scores,
we consider the partial sum:
\begin{equation}
    \Delta^2_S = \sum_{i \in S} \frac{y_i^2}{\lambda_i},
    \hspace{.5em} S \subset \{1, \ldots, N\}.
\end{equation}
We will refer to this metric as the {\bf partial Mahalanobis distance}.
Although $S$ can be any subset of $\{1, \ldots, N\}$, in this paper,
we use this metric to compare principal components with
large $S=[1, \ldots, n]$ and small $S=[n+1, \ldots, N]$ proportion of variance.
To implement this,
we simply calculate the Mahalanobis distance with features
projected to the corresponding principal components to avoid numerical instability.

\begin{table}[tb]
\center
\scriptsize
\begin{tabular}{|c|cccccc|}
\hline
Method &
\begin{tabular}[c]{c} Conditional \\ (original) \end{tabular} & 
\begin{tabular}[c]{c} Conditional \\ P(10-512) \end{tabular} & 
\begin{tabular}[c]{c} Conditional \\ P(1-9) \end{tabular} & 
\begin{tabular}[c]{c} Marginal \\ (ours) \end{tabular} & 
\begin{tabular}[c]{c} Marginal \\ P(10-512) \end{tabular} &
Euclidean \\
\hline
AUROC & {\bf 93.92} & {\bf 94.05} & 88.09 &  {\bf 93.92} & {\bf 93.94} & 89.33 \\
\hline \hline
\multirow{2}{*}{Method} &
\multicolumn{2}{c}{Conditional} & \multicolumn{2}{c}{Marginal} & \multicolumn{2}{c|}{Marginal} \\
& \multicolumn{2}{c}{(original)} & \multicolumn{2}{c}{(ours)} & \multicolumn{2}{c|}{P(10-512)} \\
\hline
Ensemble       & o & o & o & o & o & o \\
Pre-Processing &   & o &   & o &   & o \\
\hline
AUROC & 98.37 & {\bf 99.14} & 98.33 & {\bf 99.10} & 98.03 & {\bf 99.14 } \\
\hline
\end{tabular}
\vskip 1em
\caption{
AUROC for OoD detection of the partial and marginal Mahalanobis distance.
Here we use ResNet trained on CIFAR-10, and the OoD data is SVHN.
``P($i-j$)'' denotes the partial Mahalanobis distance using
the $i$-th to $j$-th principal components.
``Conditional'' and ``Marginal'' denote
the original \citep{Lee2018AAttacks} and our marginal Mahalanobis distances.
``Euclidean'' denotes the Euclidean distance from the mean.
``Ensemble'' and ``Pre-processing'' denote features ensemble
and input pre-processing proposed by \citet{Lee2018AAttacks}.}
\label{table:partial_marginal}
\end{table}
\noindent
{\bf Experiments}
We compare two partial Mahalanobis distances to analyze
behavior of the final features of anomalous inputs.
The first one is $\Delta^2_{[1, 9]} = \sum_{i=1}^9 y_i^2 / \lambda_i$ and
the second one is $\Delta^2_{[10, 512]} = \sum_{i=10}^{512} y_i^2 / \lambda_i$,
where the principal components are sorted in descending order with respect to 
the proportion of variance.
We arbitrarily chose the first $9$ principal components,
which explains about $70\%$ of variance.
Table~\ref{table:partial_marginal} shows OoD detection performance
using different confidence scores.
The results show that $\Delta^2_{[10, 512]}$ (Conditional P(10-512))
solely achieves the performance
of the original Mahalanobis confidence score
whilst using information on principal components with small explained variance.
On the contrary, the performance of
$\Delta^2_{[1, 9]}$ (Conditional P(1-9)) performs similar or worse
when compared the Euclidean distance
while using principal components explaining most of the variance in training data.
This observation suggests that the subspace spanned by
principal components with small explained variance
contains information critical for anomaly detection,
while it may not provide useful information for classification.
As Euclidean distance is not particularly sensitive to
principle components with small explained variance but Mahalanobis distance is,
the difference in results for the these distances also suggests that
this is information that is meaningful for anomaly detection.

\subsection{Marginal Mahalanobis Distance}
{\bf Definition}
Our previous analysis of the partial Mahalanobis distances suggests that
principal components with small explained variance
contribute to OoD detection,
while it may not contain
information relevant to classification.
Motivated by this observation,
we propose another variant of the Mahalanobis distance under the much more stronger assumption that
the distribution of intermediate feature is a simple Gaussian,
ignoring class information.
We call this metric the {\bf marginal Mahalanobis distance}.
Here, the model parameters are estimated as follows:
\begin{equation}
\hat \mu = \frac{1}{n} \sum_{i=1}^n f(\mv x_i), \hspace{1em}
\hat \Sigma =  \frac{1}{n} \sum_{i=1}^n (f(\mv x_i) - \hat \mu)(f(\mv x_i) - \hat \mu)^T.
\end{equation}
Based on the estimated parameters,
we use the Mahalanobis distance from $\hat \mu$, the global mean:
\begin{equation}
    M'(\mv x) = -(f(\mv x) - \hat \mu)^T \hat \Sigma^{-1} (f(\mv x) - \hat \mu)
\end{equation}
as a confidence score instead of using class conditional distributions.

\noindent
{\bf Experiments}
Table~\ref{table:partial_marginal} shows that
the marginal Mahalanobis distance,
a score which does not use class information,
achieves almost the same performance
as the original class conditional Mahalanobis score.
We also take the partial Mahalanobis distance for the marginal Mahalanobis distance.
``Marginal P(10-512)'' effectively detects OoD inputs
while only using principal components with small explained variance,
which do not contribute to classification performance
(Section~\ref{section:observation_on_intermediate_features}),
supporting our hypothesis.
We also show feature ensemble and input pre-processing proposed by \citet{Lee2018AAttacks}
are also effective on our method.

\subsection{Combine Detection Methods to Improve Detection Performance}
The original paper \citep{Lee2018AAttacks} motivates the Mahalanobis confidence score
by likening a softmax classifier to a generative classifier model 
and demonstrates competitive performance between the softmax classifer and classification 
based on the confidence score. 
Thus, it seems that the authors consider their method to be fundamentally based on 
classifier prediction confidence. 
A popular OoD detection method ODIN \citep{Hendrycks2017a} makes use of
the output of softmax function, which is directly related to classification.
This would suggest that combining their method and ODIN would not work well or reduce robustness.

However, our analysis suggests that the Mahalanobis confidence score does not use
information critical for classification.
Consequently, it is expected that the Mahalanobis confidence score and ODIN
use different information to detect OoD inputs (Table~\ref{table:classify_detection_methods}).
Motivated by this analysis, we propose combining the Mahalanobis confidence score and ODIN for OoD detection.
As the Mahalanobis confidence score uses logistic regression for feature ensemble,
we simply add a score from ODIN as a new feature.
The experiments in Section~\ref{section:experiments} shows this combined method
improves detection performance even for unseen data.

For adversarial examples detection, detection methods based on prediction confidence
may not be effective since adversarial attacks directly leverage prediction confidence.
Results by \citet{Lee2018AAttacks} support this,
showing that combining kernel density and Bayesian uncertainty estimation \citep{Feinman2017DetectingArtifacts}
often do not work well for unseen attacks.
We also provide experiments for the combination
of the Mahalanobis score and ODIN for adversarial examples in
the supplemental material, and show that is becomes more fragile to unseen data.

\section{Experiments} \label{section:experiments}

\noindent
{\bf Evaluation}
We evaluate confidence scores on threshold-based detectors of test data and anomalous inputs.
We compare the performance with the area under the receiver operating characteristic curve (AUROC).

\noindent
{\bf Settings}
For experiments on intermediate features of
ResNet \citep{He2016} and DenseNet \citep{Huang2017DenselyNetworks},
our implementation is based on code by \citet{Lee2018AAttacks}
\footnote{\url{https://github.com/pokaxpoka/deep_Mahalanobis_detector}}.
We evaluate the networks on several data sets:
CIFAR-10, CIFAR-100 \citep{Krizhevsky2009LearningImages},
SVHN \citep{Netzer2011ReadingLearning},
TinyImageNet \citep{Chrabaszcz2017ADatasets, Deng2009ImageNet:Database},
and LSUN \citep{Yu2016LSUN:Loop}.
For adversarial examples detection,
we evaluate on four types of attacks:
FGSM \citep{Goodfellow2015ExplainingAdversarial},
BIM \citep{Kurakin2017AdversarialWorld},
DeepFool \citep{Moosavi-Dezfooli2016DeepFool:Networks},
and CW \citep{Carlini2017AdversarialMethods}.
We use both input pre-processing and feature ensemble.
Details on hyperparameters are provided in the supplemental material.

\noindent
{\bf OoD Detection}
\begin{table}[tb]
\center
\scriptsize
\setlength\tabcolsep{4pt}
\begin{tabular}{|c|c|cccc|cccc|}
\hline
\multirow{3}{*}{In-dist} & \multirow{3}{*}{OOD} & 
\multicolumn{4}{c|}{Detection of test data} & \multicolumn{4}{c|}{Detection of unseen data} \\
& & ODIN & original & marginal & marginal & ODIN & original & marginal & marginal \\
& & \citep{Liang2018EnhancingNetworks} & \citep{Lee2018AAttacks} & (ours)   & + ODIN
& \citep{Liang2018EnhancingNetworks} & \citep{Lee2018AAttacks} & (ours)   & +ODIN    \\
\hline \hline
\multirow{3}{*}{\begin{tabular}[c]{c} CIFAR-10    \\ (DenseNet) \end{tabular}}
& SVHN         & 95.5 & 98.3 & 97.7 &{\bf 99.4 }& --- & --- & --- & --- \\
& TinyImagenet & 98.5 & 99.0 & 99.2 &{\bf 99.6 }& 90.3& 96.5& 92.9&{\bf 98.8 } \\
& LSUN         & 99.2 & 99.3 & 99.4 &{\bf 99.8 }& 94.1& 98.4& 93.0&{\bf 99.4 } \\
\hline
\multirow{3}{*}{\begin{tabular}[c]{c} CIFAR-100   \\ (DenseNet) \end{tabular}}
& SVHN         & 93.8 & 97.1 & 97.4 &{\bf 99.0 }& --- & --- & --- & --- \\
& TinyImagenet & 85.2 & 97.4 & 97.6 &{\bf 98.9 }& 68.2& 71.9& 74.1&{\bf 86.3 } \\
& LSUN         & 85.5 & 97.9 & 98.0 &{\bf 99.2 }& 66.3& 68.8& 71.6&{\bf 84.5 } \\
\hline
\multirow{3}{*}{\begin{tabular}[c]{c} SVHN        \\ (DenseNet) \end{tabular}}
& CIFAR-10     & 91.4 &{\bf 99.0 }& 98.5 & 98.8 & --- & --- & --- & --- \\
& TinyImagenet & 95.1 & 99.9 & 99.9 & 99.9 & 95.1& 99.9& 99.9& 99.9\\
& LSUN         & 94.5 & 99.9 & 99.9 & 99.9 & 94.5& 99.9& 99.9& 99.9\\
\hline
\multirow{3}{*}{\begin{tabular}[c]{c} CIFAR-10    \\ (ResNet) \end{tabular}}
& SVHN         & 96.7 & 99.1 & 99.1 &{\bf 99.3 }& --- & --- & --- & --- \\
& TinyImagenet & 94.0 & 99.5 & 99.5 & 99.5 & 82.9& 87.8& 89.4&{\bf 96.4 } \\
& LSUN         & 94.1 & 99.7 & 99.8 & 99.8 & 81.1& 86.1& 88.1&{\bf 96.9 } \\
\hline
\multirow{3}{*}{\begin{tabular}[c]{c} CIFAR-100   \\ (ResNet) \end{tabular}}
& SVHN         & 93.9 & 98.4 & 97.6 &{\bf 98.8 }& --- & --- & --- & --- \\
& TinyImagenet & 87.6 & 98.2 & 98.3 &{\bf 98.6 }& 57.0& 58.4&{\bf 85.1 } & 70.0\\
& LSUN         & 85.6 & 98.2 & 98.4 & 98.4 & 51.7& 43.4&{\bf 76.7 } & 56.4\\
\hline
\multirow{3}{*}{\begin{tabular}[c]{c} SVHN        \\ (ResNet) \end{tabular}}
& CIFAR-10     & 92.1 & 99.3 & 99.2 & 99.3 & --- & --- & --- & --- \\
& TinyImagenet & 92.0 & 99.9 & 99.9 & 99.9 & 92.9& 99.7& 99.7& 99.7\\
& LSUN         & 89.4 & 99.9 & 99.9 & 99.9 & 90.7& 99.9& 99.9& 99.9\\
\hline
\end{tabular}
\vskip .1in
\caption{AUROC for OoD detection on neural classifiers using
the marginal Mahalanobis confidence scores (marginal), 
the Mahalanobis confidence scores by \citet{Lee2018AAttacks} (original),
and ODIN \citep{Liang2018EnhancingNetworks}.
Our marginal confidence score which does not use class information
performs almost as well as the original scores which are based on class conditional distributions.
Also, we combine the marginal Mahalanobis confidence score and ODIN,
and show it improves detection performance even for unseen data.}
\label{table:OoDdetection}
\vskip -0.1in
\end{table}
Table~\ref{table:OoDdetection} shows
performance of OoD detection on intermediate features of
neural classifiers.
``Detection of test data'' is the OoD detectors trained on the corresponding OoD data,
and ``Detection of unseen data'' is evaluation of the robustness,
which shows detection performance of detectors trained on 
the OoD inputs in the first row (SVHN or CIFAR-10) for unseen data.
In both settings,
OoD detection with our marginal Mahalanobis confidence score
performs as well as the original score by \citet{Lee2018AAttacks},
which verifies our hypothesis.
Based on the result of our analysis,
we combine the marginal Mahalanobis distance and ODIN \citep{Liang2018EnhancingNetworks}.
This method improves detection performance in most cases even for unseen data detection.
The combined method is also as effective when the original Mahalanobis score is used
(the results are provided in the supplemental material).

\noindent
{\bf Adversarial Examples Detection}
\begin{table}[tb]
\center
\scriptsize
\begin{tabular}{|c|c|c|cccc|ccc|}
\hline
\multirow{2}{*}{Model} & \multirow{2}{*}{Dataset} & \multirow{2}{*}{Score} & 
\multicolumn{4}{c|}{Detection of known attack} & \multicolumn{3}{c|}{Detection of unknown attack} \\
 &  &  & FGSM & BIM & DeepFool & CW & BIM & DeepFool & CW   \\
\hline \hline
\multirow{12}{*}{DenseNet} 
& \multirow{4}{*}{CIFAR-10    } 
  & LID             &      98.3 &      99.7 &{\bf  85.0}&      80.3 &      94.0 &      70.4 &      68.8 \\
& & Original        &      99.9 &      99.8 &      83.4 &{\bf  84.9}&{\bf  99.5}&{\bf  83.2}&{\bf  85.7}\\
& & Marginal (ours) &      99.9 &      99.8 &      81.1 &      82.3 &      99.2 &      79.3 &      81.2 \\
& & Marginal P(10-) &      99.9 &      99.6 &      82.0 &      83.9 &      99.3 &      80.6 &      83.0 \\
\cline{2-10}
& \multirow{4}{*}{CIFAR-100   } 
  & LID             &      99.3 &      98.1 &      69.8 &      69.6 &      78.7 &      68.2 &      68.8 \\
& & Original        &      99.9 &{\bf  99.3}&{\bf  77.6}&{\bf  82.7}&{\bf  98.7}&{\bf  77.0}&{\bf  80.2}\\
& & Marginal (ours) &      99.9 &      99.2 &      76.9 &      81.6 &      98.1 &      74.3 &      78.7 \\
& & Marginal P(10-) &      99.9 &      99.1 &      76.7 &      81.8 &      98.3 &      74.5 &      79.3 \\
\cline{2-10}
& \multirow{4}{*}{SVHN        } 
  & LID             &      99.3 &      94.8 &      91.9 &      94.1 &      92.0 &      80.1 &      81.2 \\
& & Original        &      99.8 &{\bf  99.3}&{\bf  95.1}&{\bf  96.5}&{\bf  99.2}&{\bf  94.2}&{\bf  96.4}\\
& & Marginal (ours) &      99.9 &      99.0 &      92.2 &      95.0 &      98.8 &      90.2 &      93.4 \\
& & Marginal P(10-) &      99.9 &      99.1 &      92.9 &      95.4 &      99.0 &      91.1 &      95.0 \\
\hline
\multirow{12}{*}{ResNet} 
& \multirow{4}{*}{CIFAR-10    } 
  & LID             &      99.7 &      96.6 &      88.5 &      83.0 &      94.0 &      73.6 &      78.1 \\
& & Original        &     100.0 &      99.6 &{\bf  91.4}&{\bf  95.9}&      99.0 &      80.6 &      94.3 \\
& & Marginal (ours) &     100.0 &      99.6 &      91.0 &      95.8 &{\bf  99.5}&{\bf  86.2}&{\bf  95.5}\\
& & Marginal P(10-) &     100.0 &      99.5 &      90.9 &      95.3 &      99.1 &      82.0 &      94.5 \\
\cline{2-10}
& \multirow{4}{*}{CIFAR-100   } 
  & LID             &      98.4 &      97.0 &      71.8 &      78.4 &      59.8 &      65.1 &      76.2 \\
& & Original        &      99.8 &      96.7 &{\bf  85.3}&{\bf  92.0}&      96.4 &{\bf  81.9}&{\bf  91.0}\\
& & Marginal (ours) &      99.8 &      97.7 &      73.5 &      90.4 &      97.1 &      69.8 &      87.5 \\
& & Marginal P(10-) &      99.8 &      97.7 &      73.5 &      90.2 &{\bf  97.6}&      70.5 &      88.6 \\
\cline{2-10}
& \multirow{4}{*}{SVHN        } 
  & LID             &      97.8 &      90.6 &      92.2 &      88.3 &      82.2 &      68.5 &      75.1 \\
& & Original        &      99.6 &      97.1 &{\bf  95.7}&{\bf  92.2}&      95.8 &      73.9 &      87.9 \\
& & Marginal (ours) &      99.6 &{\bf  97.2}&      95.2 &      91.9 &      95.3 &      70.2 &      86.2 \\
& & Marginal P(10-) &      99.6 &      97.0 &      95.2 &      91.9 &{\bf  95.9}&{\bf  75.6}&{\bf  88.0}\\
\hline
\end{tabular}
\vskip .1in
\caption{AUROC for adversarial examples detection on neural classifiers using
LID  \citep{Ma2018CharacterizingDimensionality},
the class conditional Mahalanobis distance-based score by \citet{Lee2018AAttacks} (Original),
and our marginal Mahalanobis distance-based score (Marginal).
``Marginal P(10-)'' denotes the partial Mahalanobis distance using 10-th to 512-th principal components.
Our confidence score achieves competitive performance when compared to the original method.}
\label{table:adversarialdetection}
\end{table}
Table~\ref{table:adversarialdetection} shows adversarial examples detection performance.
Our Mahalanobis confidence score
is as effective as the class conditional method by \citet{Lee2018AAttacks}.
``Detection of unknown attack'' evaluates a detector trained on FGSM,
and shows that our confidence score also generalizes well to other types of attacks.
The partial Mahalanobis score seems to improve robustness,
but does not always outperforms all other methods.
However, as our marginal and partial Mahalanobis score works well,
these results verify our hypothesis that the Mahalanobis score does not depend on prediction confidence
for classification task.

\section{Conclusion}
The Mahalanobis distance-based confidence score \citep{Lee2018AAttacks},
an anomaly detection method for pre-trained neural classifiers,
achieves state-of-the-art performance both in OoD and adversarial examples detection.
This paper is the first work analyzing why this method can detect anomalous inputs effectively
while imposing the implausible assumption that class conditional distributions
of intermediate features have tied covariance,
and shows that the reason for its effectiveness has been misunderstood.
The motivation for the Mahalanobis confidence score was based on prediction confidence for classification,
but we demonstrate that it utilizes information that is not critical for classification task
to distinguish between in-distribution and anomalous data.
Our analysis suggests that the Mahalanobis confidence score
makes use of properties of OoD inputs that is not utilized in
other OoD detection method such as ODIN \citep{Liang2018EnhancingNetworks},
so we propose combining these two methods and show this idea improves
detection performance and robustness.
This work provides critical insight into
the new standard anomaly detection method and
the behavior of neural networks towards anomalous inputs.
 
\ifsubmission
\section*{Acknowledgements}
This paper has benefited from advice and English language editing from
Masayuki Takeda and Toshinori Kitamura.
This work was supported by JSPS KAKENHI (JP19K03642, JP19K00912) and RIKEN AIP Japan.
\fi

\bibliography{references}

\clearpage

\ifsubmission
\appendix
\section{Background}
We briefly explain ODIN \citep{Liang2018EnhancingNetworks}
and the Mahalanobis distance-based confidence score \citep{Lee2018AAttacks}.

\subsection{ODIN: Out-of-DIstribution detector for Neural networks}
ODIN \citep{Liang2018EnhancingNetworks} is a simple yet effective improvement
of a baseline OoD method on pre-trained neural classifiers by \citet{Hendrycks2017a},
which uses the maximum output from softmax function as a confidence score.
ODIN is built on two components: temperature scaling and input preprocessing.

\noindent
{\bf Temperature Scaling}
Let $f=(f_1, \ldots, f_N)$ be a pre-trained neural classifier for $N$ classes.
The temperature scaling is a variant of softmax function:
\begin{equation}
    S_i(x; T) = \frac{\exp(f_i(x)/T)}{\sum_{j=1}^N \exp(f_j(x)/T)},
\end{equation}
where $T \in \mathbb R^+$ is the temperature scaling parameter.
The standard softmax function $T=1$ is used during training.

\noindent
{\bf Input Pre-processing}
ODIN applies pre-processing to the input image as follows:
\begin{equation}
\tilde x = x - \varepsilon \mathrm{sign} (-\nabla_x \log S_{\hat y}(x; T))
\end{equation}
where $\varepsilon$ is the magnitude parameter.
The formulation of this input pre-processing is inspired by
FGSM \citep{Goodfellow2015ExplainingAdversarial},
a popular adversarial attack.

\subsection{Mahalanobis Distance-Based Confidence Score}
\citet{Lee2018AAttacks} proposed an input pre-processing similar to that in ODIN
for their Mahalanobis confidence score.
They applies pre-processing to the input image as follows:
\begin{equation}
\tilde x = x - \varepsilon \mathrm{sign} (-\nabla_x M(x))
\end{equation}
where $M(x)$ denotes the Mahalanobis confidence score,
and $\varepsilon$ is the magnitude parameter.

\section{Hyperparameters}
The Mahalanobis distance-based confidence score and ODIN have hyperparameters.
The temperature $T$ of ODIN is chosen from $\{1, 10, 100, 1000\}$,
and the magnitude of input pre-processing for the Mahalanobis confidence score and ODIN is chosen from
$\{0, 0.0005, 0.001, \allowbreak 0.0014, 0.002, 0.0024, 0.005, 0.01, 0.05, 0.1, 0.2\}$.

For data splitting, we use the same configuration as the original implementation \citep{Lee2018AAttacks}.
The training data of the data sets are used to calculate the Mahalanobis parameters.
For OoD detection, the test data (of the original data sets)
is split into training data, validation data, and test data
for logistic regression and hyperparameter tuning.
The size of training and validation data for logistic regression is 1,000 each,
and the rest of data is used as test data.
For adversarial examples detection,
the test data is equally split into training data, adversarial examples, and noisy data.
The noisy data is images with random noise added,
which is regarded as a in-distribution data.
The size of training and validation data for logistic regression is $5\%$ each,
and the rest of data is used as test data.
For the detailed configuration, please refer to the original or our implementation.

\section{Additional Results}
We provide additional experimental results for OoD and adversarial examples detection.

\subsection{OoD Detection}
\begin{table}[!t]
\center
\scriptsize
\setlength\tabcolsep{4pt}
\begin{tabular}{|c|c|cccc|cccc|}
\hline
\multirow{3}{*}{In-dist} & \multirow{3}{*}{OOD} & 
\multicolumn{4}{c|}{Detection of test data} & \multicolumn{4}{c|}{Detection of unseen data} \\
& & original & marginal & original & marginal  & original & marginal & original & marginal \\
& & \citep{Lee2018AAttacks} & (ours)   & + ODIN & + ODIN
& \citep{Lee2018AAttacks} & (ours)   & +ODIN & +ODIN \\
\hline \hline
\multirow{3}{*}{\begin{tabular}[c]{c} CIFAR-10    \\ (DenseNet) \end{tabular}}
& SVHN         & 98.3 & 97.7 & 99.4 & 99.4 & --- & --- & --- & --- \\
& TinyImagenet & 99.0 & 99.2 & 99.6 & 99.6 & 96.5& 92.9&{\bf 98.9 } &{\bf 98.8}\\
& LSUN         & 99.3 & 99.4 & 99.8 & 99.8 & 98.4& 93.0&{\bf 99.5 } &{\bf 99.4}\\
\hline
\multirow{3}{*}{\begin{tabular}[c]{c} CIFAR-100   \\ (DenseNet) \end{tabular}}
& SVHN         & 97.1 & 97.4 & 98.8 &{\bf 99.0 }& --- & --- & --- & --- \\
& TinyImagenet & 97.4 & 97.6 & 98.9 & 98.9 & 71.9& 74.1&{\bf 86.4 } &{\bf 86.3}\\
& LSUN         & 97.9 & 98.0 & 98.9 &{\bf 99.2 }& 68.8& 71.6&{\bf 84.6 } &{\bf  84.5}\\
\hline
\multirow{3}{*}{\begin{tabular}[c]{c} SVHN        \\ (DenseNet) \end{tabular}}
& CIFAR-10     & 99.0 & 98.5 & 99.0 & 98.8 & --- & --- & --- & --- \\
& TinyImagenet & 99.9 & 99.9 & 99.9 & 99.9 & 99.9& 99.9& 99.9& 99.9\\
& LSUN         & 99.9 & 99.9 & 99.9 & 99.9 & 99.9& 99.9& 99.9& 99.9\\
\hline
\multirow{3}{*}{\begin{tabular}[c]{c} CIFAR-10    \\ (ResNet) \end{tabular}}
& SVHN         & 99.1 & 99.1 &{\bf 99.4 }& 99.3 & --- & --- & --- & --- \\
& TinyImagenet & 99.5 & 99.5 & 99.5 & 99.5 & 87.8& 89.4&{\bf 98.3 } & 96.4\\
& LSUN         & 99.7 & 99.8 & 99.7 & 99.8 & 86.1& 88.1&{\bf 98.4 } & 96.9\\
\hline
\multirow{3}{*}{\begin{tabular}[c]{c} CIFAR-100   \\ (ResNet) \end{tabular}}
& SVHN         & 98.4 & 97.6 & 98.8 & 98.8 & --- & --- & --- & --- \\
& TinyImagenet & 98.2 & 98.3 & 98.1 &{\bf 98.6 }& 58.4&{\bf 85.1 } & 65.5& 70.0\\
& LSUN         & 98.2 & 98.4 & 98.2 & 98.4 & 43.4&{\bf 76.7 } & 51.9& 56.4\\
\hline
\multirow{3}{*}{\begin{tabular}[c]{c} SVHN        \\ (ResNet) \end{tabular}}
& CIFAR-10     & 99.3 & 99.2 & 99.3 & 99.3 & --- & --- & --- & --- \\
& TinyImagenet & 99.9 & 99.9 & 99.9 & 99.9 & 99.7& 99.7& 99.7& 99.7\\
& LSUN         & 99.9 & 99.9 & 99.9 & 99.9 & 99.9& 99.9& 99.9& 99.9\\
\hline
\end{tabular}
\vskip .1in
\caption{AUROC for OoD detection on neural classifiers using
the marginal Mahalanobis confidence scores (marginal)
and the Mahalanobis confidence scores by \citet{Lee2018AAttacks}
(original).
Here, we also show the results for the combined method of the original Mahalanobis distance and ODIN.
It is also as effective as the combined method of the marginal Mahalanobis distance and ODIN.
}
\label{table:ood_odin_appendix}
\vskip -0.1in
\center
\scriptsize
\setlength\tabcolsep{4pt}
\begin{tabular}{|c|c|cccc|cccc|}
\hline
\multirow{3}{*}{In-dist} & \multirow{3}{*}{OOD} & 
\multicolumn{4}{c|}{Detection of unseen data} & \multicolumn{4}{c|}{Detection of unseen data} \\
& & original & marginal & original & marginal  & original & marginal & original & marginal \\
& & \citep{Lee2018AAttacks} & (ours)   & + ODIN & + ODIN
& \citep{Lee2018AAttacks} & (ours)   & +ODIN & +ODIN \\
\hline \hline
\multirow{3}{*}{\begin{tabular}[c]{c} CIFAR-10    \\ (DenseNet) \end{tabular}}
& SVHN         & 97.8 & 90.3 &{\bf 99.2 }& 98.2 & 97.9& 78.8&{\bf 99.3 } & 96.7\\
& TinyImagenet & --- & --- & --- & --- & 98.9& 99.0& 99.7& 99.7\\
& LSUN         & 99.3 & 99.5 & 99.7 &{\bf 99.8 }& --- & --- & --- & --- \\
\hline
\multirow{3}{*}{\begin{tabular}[c]{c} CIFAR-100   \\ (DenseNet) \end{tabular}}
& SVHN         & 91.6 & 91.5 &{\bf 95.8 }& 95.7 & 92.0& 93.0& 96.0&{\bf 96.7 } \\
& TinyImagenet & --- & --- & --- & --- & 97.5& 97.3& 98.8&{\bf 99.1 } \\
& LSUN         & 97.9 & 98.0 & 98.9 &{\bf 99.0 }& --- & --- & --- & --- \\
\hline
\multirow{3}{*}{\begin{tabular}[c]{c} SVHN        \\ (DenseNet) \end{tabular}}
& CIFAR-10     & 98.7 & 98.3 & 98.7 & 98.3 &{\bf 98.6 } & 98.0& 98.5& 98.4\\
& TinyImagenet & --- & --- & --- & --- & 99.9& 99.8& 99.9& 99.9\\
& LSUN         & 99.9 & 99.9 & 99.9 & 99.9 & --- & --- & --- & --- \\
\hline
\multirow{3}{*}{\begin{tabular}[c]{c} CIFAR-10    \\ (ResNet) \end{tabular}}
& SVHN         & 96.3 & 94.2 & 96.3 & 94.2 & 96.1& 93.2& 96.1& 93.8\\
& TinyImagenet & --- & --- & --- & --- & 99.4& 99.4& 99.4& 99.4\\
& LSUN         & 99.7 & 99.7 & 99.7 & 99.7 & --- & --- & --- & --- \\
\hline
\multirow{3}{*}{\begin{tabular}[c]{c} CIFAR-100   \\ (ResNet) \end{tabular}}
& SVHN         & 90.9 & 92.1 &{\bf 92.7 }& 89.9 & 92.5& 90.1&{\bf 92.6 } & 90.2\\
& TinyImagenet & --- & --- & --- & --- & 98.0& 98.0& 98.0& 98.0\\
& LSUN         & 98.4 & 98.6 & 98.3 &{\bf 98.7 }& --- & --- & --- & --- \\
\hline
\multirow{3}{*}{\begin{tabular}[c]{c} SVHN        \\ (ResNet) \end{tabular}}
& CIFAR-10     & 99.3 & 99.3 & 99.3 & 99.3 & 98.6& 98.4& 98.6& 98.4\\
& TinyImagenet & --- & --- & --- & --- & 99.4& 99.4& 99.4& 99.4\\
& LSUN         & 99.9 & 99.9 & 99.9 & 99.9 & --- & --- & --- & --- \\
\hline
\end{tabular}
\vskip .1in
\caption{
Unseen data detection performance
of detectors trained on the second and third OoD data.
}
\label{table:ood_odin_appendix2}
\vskip -0.1in
\end{table}
\begin{table}[!t]
\center
\scriptsize
\setlength\tabcolsep{4pt}
\begin{tabular}{|c|c|cccc|cccc|}
\hline
\multirow{3}{*}{In-dist} & \multirow{3}{*}{OOD} & 
\multicolumn{4}{c|}{Detection of test data} & \multicolumn{4}{c|}{Trained on FGSM} \\
& & original & marginal & original & marginal  & original & marginal & original & marginal \\
& & \citep{Lee2018AAttacks} & (ours)   & + ODIN & + ODIN
& \citep{Lee2018AAttacks} & (ours)   & +ODIN & +ODIN \\
\hline \hline
\multirow{3}{*}{\begin{tabular}[c]{c} CIFAR-10    \\ (DenseNet) \end{tabular}}
& SVHN         & 98.3 & 97.7 & 99.4 & 99.4 & 97.5& 96.1& 97.5& 96.1\\
& TinyImagenet & 99.0 & 99.2 & 99.6 & 99.6 & 98.8& 98.6& 98.8& 98.6\\
& LSUN         & 99.3 & 99.4 & 99.8 & 99.8 & 99.2& 99.1& 99.2& 99.1\\
\hline
\multirow{3}{*}{\begin{tabular}[c]{c} CIFAR-100   \\ (DenseNet) \end{tabular}}
& SVHN         & 97.1 & 97.4 & 98.8 &{\bf 99.0 }& 89.5& 87.0& 94.6&{\bf 96.6 } \\
& TinyImagenet & 97.4 & 97.6 & 98.9 & 98.9 & 97.3& 97.2&{\bf 97.9 } & 94.1\\
& LSUN         & 97.9 & 98.0 & 98.9 &{\bf 99.2 }& 98.0& 97.8& 98.0& 92.6\\
\hline
\multirow{3}{*}{\begin{tabular}[c]{c} SVHN        \\ (DenseNet) \end{tabular}}
& CIFAR-10     & 99.0 & 98.5 & 99.0 & 98.8 &{\bf 98.7 } & 93.7& 89.4& 84.0\\
& TinyImagenet & 99.9 & 99.9 & 99.9 & 99.9 &{\bf 99.9 } & 99.5& 99.0& 98.4\\
& LSUN         & 99.9 & 99.9 & 99.9 & 99.9 &{\bf 99.9 } & 99.7& 99.5& 99.3\\
\hline
\multirow{3}{*}{\begin{tabular}[c]{c} CIFAR-10    \\ (ResNet) \end{tabular}}
& SVHN         & 99.1 & 99.1 &{\bf 99.4 }& 99.3 & 67.5&{\bf 88.4 } & 38.2& 34.7\\
& TinyImagenet & 99.5 & 99.5 & 99.5 & 99.5 & 94.9&{\bf 99.0 } & 95.5& 95.1\\
& LSUN         & 99.7 & 99.8 & 99.7 & 99.8 & 98.0&{\bf 99.6 } & 98.3& 98.0\\
\hline
\multirow{3}{*}{\begin{tabular}[c]{c} CIFAR-100   \\ (ResNet) \end{tabular}}
& SVHN         & 98.4 & 97.6 & 98.8 & 98.8 & 93.2& 78.4&{\bf 93.3 } & 87.7\\
& TinyImagenet & 98.2 & 98.3 & 98.1 &{\bf 98.6 }&{\bf 76.3 } & 72.2& 76.2& 74.9\\
& LSUN         & 98.2 & 98.4 & 98.2 & 98.4 &{\bf 65.4 } & 61.4& 65.3& 64.1\\
\hline
\multirow{3}{*}{\begin{tabular}[c]{c} SVHN        \\ (ResNet) \end{tabular}}
& CIFAR-10     & 99.3 & 99.2 & 99.3 & 99.3 & 97.8& 97.4& 97.8& 97.5\\
& TinyImagenet & 99.9 & 99.9 & 99.9 & 99.9 & 99.3& 99.3& 99.3& 99.3\\
& LSUN         & 99.9 & 99.9 & 99.9 & 99.9 & 99.9& 99.8& 99.9& 99.8\\
\hline
\end{tabular}
\vskip .1in
\caption{AUROC for OoD detection on neural classifiers using
the marginal Mahalanobis confidence scores (marginal)
and the Mahalanobis confidence scores by \citet{Lee2018AAttacks}
(original).
We evaluate detectors trained on adversarial examples detection 
for FSGM \citep{Goodfellow2015ExplainingAdversarial}
as this setting was evaluated in \citet{Lee2018AAttacks}.
In some cases, the combined methods are worse in this setting.
This result is understandable because ODIN is directly based on the prediction confidence
acquired from the output of softmax function, which is attacked by adversarial attacks.
}
\label{table:ood_trained_on_adv}
\vskip -0.1in
\end{table}

Table~\ref{table:ood_odin_appendix} shows results similar to those provided in the main paper,
but we also provide the combined method of the original Mahalanobis score and ODIN.
It is as effective as the combined method of our marginal Mahalanobis score and ODIN.
Table~\ref{table:ood_odin_appendix2} shows unseen data detection performance
of detectors trained on the second and third OoD data.

Table~\ref{table:ood_trained_on_adv} shows OoD detection performances of detectors trained on
adversarial examples detection for FGSM,
which was provided in \citet{Lee2018AAttacks}.
In this setting, the combined methods of the Mahalanobis score and ODIN are worse in some cases.
This result is understandable because ODIN is directly based on the prediction confidence
acquired from the output of softmax function, which would be fooled by adversarial attacks.
Therefore, the confidence score of ODIN trained on adversarial examples detection
may not be applicable to OoD detection for unseen data.
However, as we demonstrate in the main paper,
the combined method can effectively detect unseen data when it is trained
to detect realistic OoD data.

\subsection{Adversarial Examples Detection}
\begin{table}[tb]
\center
\scriptsize
\begin{tabular}{|c|c|c|cccc|ccc|}
\hline
\multirow{2}{*}{Model} & \multirow{2}{*}{Dataset} & \multirow{2}{*}{Score} & 
\multicolumn{4}{c|}{Detection of known attack} & \multicolumn{3}{c|}{Detection of unknown attack} \\
 &  &  & FGSM & BIM & DeepFool & CW & BIM & DeepFool & CW   \\
\hline \hline
\multirow{12}{*}{DenseNet} 
& \multirow{5}{*}{CIFAR-10    } 
  & LID             &      98.3 &      99.7 &      85.0 &      80.3 &      94.0 &      70.4 &      68.8 \\
& & Original        &      99.9 &      99.8 &      83.4 &      84.9 &{\bf  99.5}&{\bf  83.2}&{\bf  85.7}\\
& & Marginal (ours) &      99.9 &      99.8 &      81.1 &      82.3 &      99.2 &      79.3 &      81.2 \\
& & Marginal + ODIN &      99.9 &{\bf 100.0}&{\bf  88.4}&      85.7 &      99.2 &      79.3 &      81.2 \\
& & Marginal + LID  &      99.9 &      99.9 &      87.9 &{\bf  89.6}&      98.7 &      76.9 &      79.5 \\
\cline{2-10}
& \multirow{5}{*}{CIFAR-100   } 
  & LID             &      99.3 &      98.1 &      69.8 &      69.6 &      78.7 &      68.2 &      68.8 \\
& & Original        &      99.9 &{\bf  99.3}&      77.6 &      82.7 &{\bf  98.7}&      77.0 &{\bf  80.2}\\
& & Marginal (ours) &      99.9 &      99.2 &      76.9 &      81.6 &      98.1 &      74.3 &      78.7 \\
& & Marginal + ODIN &      99.6 &      99.2 &{\bf  84.4}&      83.4 &       9.0 &{\bf  78.9}&      65.7 \\
& & Marginal + LID  &      99.8 &      99.2 &      79.7 &{\bf  84.8}&      63.6 &      75.2 &      75.5 \\
\cline{2-10}
& \multirow{5}{*}{SVHN        } 
  & LID             &      99.3 &      94.8 &      91.9 &      94.1 &      92.0 &      80.1 &      81.2 \\
& & Original        &      99.8 &      99.3 &      95.1 &      96.5 &{\bf  99.2}&{\bf  94.2}&{\bf  96.4}\\
& & Marginal (ours) &      99.9 &      99.0 &      92.2 &      95.0 &      98.8 &      90.2 &      93.4 \\
& & Marginal + ODIN &      99.8 &      99.3 &{\bf  95.6}&{\bf  97.7}&      94.0 &      80.9 &      78.5 \\
& & Marginal + LID  &      99.9 &      99.3 &      95.0 &      97.6 &      98.7 &      89.8 &      94.4 \\
\hline
\multirow{12}{*}{ResNet} 
& \multirow{5}{*}{CIFAR-10    } 
  & LID             &      99.7 &      96.6 &      88.5 &      83.0 &      94.0 &      73.6 &      78.1 \\
& & Original        &     100.0 &      99.6 &      91.4 &      95.9 &      99.0 &      80.6 &      94.3 \\
& & Marginal (ours) &     100.0 &      99.6 &      91.0 &      95.8 &{\bf  99.5}&{\bf  86.2}&{\bf  95.5}\\
& & Marginal + ODIN &      99.9 &{\bf  99.9}&{\bf  92.0}&      96.0 &      98.2 &      79.7 &      91.6 \\
& & Marginal + LID  &     100.0 &      99.7 &      91.7 &{\bf  96.1}&      97.4 &      82.8 &      89.3 \\
\cline{2-10}
& \multirow{5}{*}{CIFAR-100   } 
  & LID             &      98.4 &      97.0 &      71.8 &      78.4 &      59.8 &      65.1 &      76.2 \\
& & Original        &      99.8 &      96.7 &      85.3 &{\bf  92.0}&      96.4 &{\bf  81.9}&{\bf  91.0}\\
& & Marginal (ours) &      99.8 &      97.7 &      73.5 &      90.4 &      97.1 &      69.8 &      87.5 \\
& & Marginal + ODIN &      99.8 &      98.8 &{\bf  87.7}&      91.9 &{\bf  97.2}&      70.9 &      89.5 \\
& & Marginal + LID  &      99.7 &      98.8 &      78.8 &      91.8 &      89.8 &      64.5 &      78.7 \\
\cline{2-10}
& \multirow{5}{*}{SVHN        } 
  & LID             &      97.8 &      90.6 &      92.2 &      88.3 &      82.2 &      68.5 &      75.1 \\
& & Original        &      99.6 &      97.1 &      95.7 &      92.2 &      95.8 &{\bf  73.9}&      87.9 \\
& & Marginal (ours) &      99.6 &      97.2 &      95.2 &      91.9 &      95.3 &      70.2 &      86.2 \\
& & Marginal + ODIN &      99.6 &      97.8 &      95.7 &      92.0 &      95.6 &      69.5 &      87.2 \\
& & Marginal + LID  &{\bf  99.7}&      97.8 &{\bf  95.8}&{\bf  93.6}&{\bf  96.0}&      70.2 &{\bf  88.6}\\
\hline
\end{tabular}
\vskip .1in
\caption{AUROC for adversarial example detection on neural classifiers using
LID  \citep{Ma2018CharacterizingDimensionality},
the class conditional Mahalanobis distance-based score by \citet{Lee2018AAttacks} (Original),
and our marginal Mahalanobis distance-based score (Marginal).
Here, we also evaluate combined methods,
but they sometimes do not work for unseen attacks.}
\label{table:additional_adversarial}
\end{table}
We observe that ODIN and LID harm robustness of the model in some cases
when they are combined with the Mahalanobis score.
For ODIN, this result is understandable because adversarial attacks directly leverage
the prediction confidence, which is utilized by ODIN as explained above.
Therefore, ODIN may not effectively detect unseen attacks and harms the detection performance.

\section{Applicability and Limitations}
\citet{Lee2018AAttacks} has suggested that 
their confidence score
may be applied to other architectures
such as few-shot learning.
Our experiments show that the situation is not that simple;
the method does not always work effectively on metric learning.
This observation suggests that the Mahalanobis distance-based confidence score
utilizes a unique property
of ordinary neural classifiers,
so caution is needed when applying the method to other architectures.

\noindent
{\bf Settings}
We use the implementation by \citet{Chen2019AClassification}
\footnote{\url{https://github.com/wyharveychen/CloserLookFewShot}}.
We evaluate two types of distance-based metric learning:
Prototypical Networks \citep{Snell2017PrototypicalLearning} and
Matching Networks \citep{Vinyals2016MatchingLearning}
with four-layer convolutional networks and a simplified ResNet
(Conv-4 and ResNet-10 in \citet{Chen2019AClassification}).
In this experiment, we use two pairs of data sets:
CUB-200-2011 \citep{Wah2011TheDataset} vs.
mini-ImageNet \citep{Vinyals2016MatchingLearning, Chen2019AClassification, Ravi2017OptimizationLearning},
and Omniglot vs.\ EMNIST \citep{Cohen2017EMNIST:Letters}.
For both of the pairs,
the models are trained on ``base'' data,
and anomaly detection is evaluated on ``novel'' data of training and OoD data
How the data sets are split are explained in \citet{Chen2019AClassification}.
For comparison, we evaluate neural classifiers using the same architectures
and cross entropy as a loss function.

\begin{table}[tb]
\center
\scriptsize
\begin{tabular}{cccccccc}
\hline
Model & Method & Score & in-dist & OoD &
\begin{tabular}[c]{c} TNR at\\ TPR 95\% \end{tabular} & AUROC& 
\begin{tabular}[c]{c} Detection\\ Accuracy\end{tabular} \\
\hline \hline
\multirow{6}{*}{Conv}
& (Classifier) & Maha & \multirow{6}{*}{Omniglot} & \multirow{6}{*}{EMNIST}
& 98.27 & {\bf 80.65} & 75.65 \\
& (Classifier) & Eucl &  &  & 95.77 & 47.48 & 58.83  \\
& ProtoNet     & Maha &  &  & 97.80 & 80.34 & 72.84  \\
& ProtoNet     & Eucl &  &  & 98.05 & {\bf 84.71} & 76.85  \\
& MatchNet     & Maha &  &  & 96.02 & 57.47 & 60.04  \\
& MatchNet     & Eucl &  &  & 98.35 & {\bf 72.43} & 75.09  \\
\hline
\multirow{6}{*}{Conv}
& (Classifier) & Maha & \multirow{6}{*}{CUB} & \multirow{6}{*}{ImageNet} 
&  9.34 & {\bf 60.21} & 57.69  \\
& (Classifier) & Eucl &  &  &  8.57 & 52.73 & 52.38  \\
& ProtoNet     & Maha &  &  & 32.93 & {\bf 61.97} & 58.92 \\
& ProtoNet     & Eucl &  &  & 16.22 & 55.68 & 54.94 \\
& MatchNet     & Maha &  &  & 25.31 & {\bf 64.07} & 61.69 \\
& MatchNet     & Eucl &  &  & 12.50 & 53.46 & 54.17 \\
\hline
\multirow{6}{*}{ResNet}
& (Classifier) & Maha & \multirow{6}{*}{CUB} & \multirow{6}{*}{ImageNet} 
& 14.10 & {\bf 51.60} & 54.20 \\
& (Classifier) & Eucl &  &  & 20.87 & 50.50 & 51.01 \\
& ProtoNet     & Maha &  &  & 21.07 & {\bf 61.93} & 59.99 \\
& ProtoNet     & Eucl &  &  & 19.58 & 46.49 & 50.10 \\
& MatchNet     & Maha &  &  & 35.51 & {\bf 67.59} & 63.67 \\
& MatchNet     & Eucl &  &  & 21.36 & 52.07 & 52.43 \\
\hline
\end{tabular}
\vskip 1em
\caption{Marginal Mahalanobis distance-based OoD detection on distance-based metric learning models.
We evaluate Prototypical Networks (ProtoNet), Matching Networks (MatchNet),
and ordinary neural classifiers (Classifier).
``Conv'' and ``ResNet'' denote Conv-4 and ResNet-10 in \citet{Chen2019AClassification}.
``Maha'' and ``Eucl'' denote the Mahalanobis and Euclidean distance.
\label{table:distancebased}}
\end{table}

\noindent
{\bf Results}
Table~\ref{table:distancebased} shows OoD detection performance of 
the Mahalanobis distance-based confidence score
evaluated on distance-based metric learning architectures.
We only use the final features,
without using feature ensembles nor input pre-processing.
Since in-distribution data sets evaluated here have too many labels,
we only evaluate our marginal method.
For CUB vs.\ mini-ImageNet,
it seems that neural classifiers and metric learning models
exhibit similar OoD behavior on the final features;
the Mahalanobis distance outperforms the Euclidean distance.
However, metric learning models exhibit unexpected behaviors for
Omniglot vs.\ EMNIST;
the detection performance of the Euclidean distance is better
on the metric learning models for this pair of data sets.
Our experimental results suggest that the intermediate features of distance-based metric learning models
do not always have the property that
principal components with small explained variance provide information for anomaly detection.
This observation suggests that the Mahalanobis distance-based confidence score
utilizes the unique property of ordinary neural classifiers,
and its performance may depend on model architecture along with the choice of
in-distribution and OoD data.
Furthermore, for CUB vs.\ ImageNet,
Table~\ref{table:distancebased} shows that detection performance on 
the neural classifier and Matching Networks
is influenced by model architecture.
In conclusion, caution is required when
applying the Mahalanobis distance-based confidence score
to new architectures or data sets.

\section{Analysis of Input Pre-processing}
\citet{Lee2018AAttacks} has proposed the following input pre-processing:
\begin{equation}
    \hat{\mathbf x} = \mathbf x + \varepsilon \mathrm{sign}(\nabla_{\mathbf x} M(\mathbf x))
\end{equation}
where $\varepsilon$ is the magnitude of noise.
While this considerably improves the detection performance,
the mechanism of its contribution has not been analyzed well.
This method is similar to the pre-processing by \citet{Liang2018EnhancingNetworks}
that takes the gradient of the cross-entropy.
For cross-entropy loss,
\citet{Liang2018EnhancingNetworks} demonstrated that the norms of the gradients for in-distribution data
tend to be larger than those for OoD data,
so the pre-processing increases confidence for in-distribution data more than
that for OoD data.
However, in the input pre-processing for the Mahalanobis distance,
we observe the opposite behavior.
Figure~\ref{fig:resnet_cifar10_svhn} (a) shows that L1 norms of the differences between
the original and pre-processed features on the penultimate layer
for the class conditional Mahalanobis distance.
It shows that L1 norms of SVHN (OoD data) tend to be larger than
those for CIFAR-10 (in-distribution data).
As a result, the input pre-processing lowers the Mahalanobis distance and
increase confidence for OoD inputs
more than for in-distribution data.
However, as shown in Figure~\ref{fig:resnet_cifar10_svhn} (b) and (c),
the mean Mahalanobis distance on the final features of SVHN increase much faster than
that of CIFAR-10 as the magnitude increases after it reaches the minimum value.
For $\varepsilon=0.01$,
the mean Mahalanobis distance of CIFAR-10 is still lower than that for $\varepsilon=0$,
while that of SVHN is much larger than the original value.
As a result, the detection performance is improved from $98.37$ to $99.14$ in AUROC.
These results and Figure~\ref{fig:resnet_cifar10_svhn} (d) suggest that
a proper magnitude that increases the Mahalanobis distances
of OoD inputs more than those of in-distribution data,
the detection performance will be improved.

We provide further experimental results analyzing input pre-processing by \citet{Lee2018AAttacks}.
Figure~\ref{fig:resnet_cifar10_svhn} to \ref{fig:densenet_svhn_lsun_resize}
show that input pre-processing does not always improve detection performance.
When L1 norms of input pre-processing for
OoD inputs on the final features are sufficiently larger
than these for in-distribution inputs,
detection performance will be improved
(Figure~\ref{fig:resnet_cifar10_svhn}, \ref{fig:resnet_cifar100_svhn}, \ref{fig:densenet_cifar100_svhn}).
However, in some cases,
input pre-processing does not improve performance
as it does not make the Mahalanobis distance of OoD inputs sufficiently large.
When L1 norms of in-distribution inputs are larger than these of OoD inputs,
i.e.\ the situation corresponding to what is discussed in \citet{Liang2018EnhancingNetworks},
detection performance does not improve
(Figure~\ref{fig:densenet_cifar10_imagenet_resize}, \ref{fig:densenet_cifar10_lsun_resize},
\ref{fig:densenet_cifar100_imagenet_resize}, \ref{fig:densenet_cifar100_lsun_resize}).
To sum up, our experiments show that the effect of input pre-processing depends on
the network architecture and choice of data set.
However, we demonstrate that the detection performance is improved
when OoD inputs on the final features becomes larger
than these for in-distribution inputs
which is different from previous discussion for input pre-processing for ODIN.
\begin{figure}[tb]
\center
\begin{tabular}{cccc}
\bmvaHangBox{\includegraphics[height=1.in]{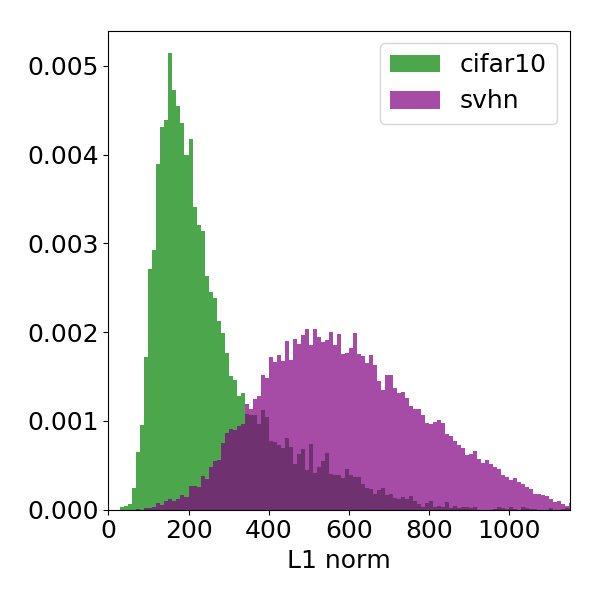}} &
\bmvaHangBox{\includegraphics[height=1.in]{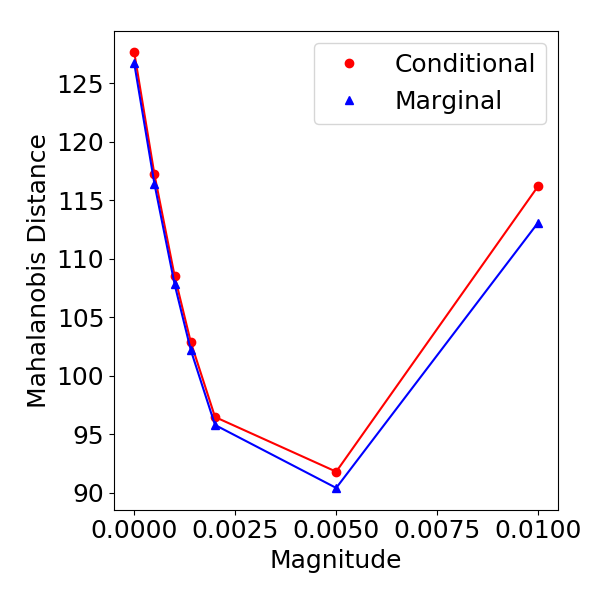}} &
\bmvaHangBox{\includegraphics[height=1.in]{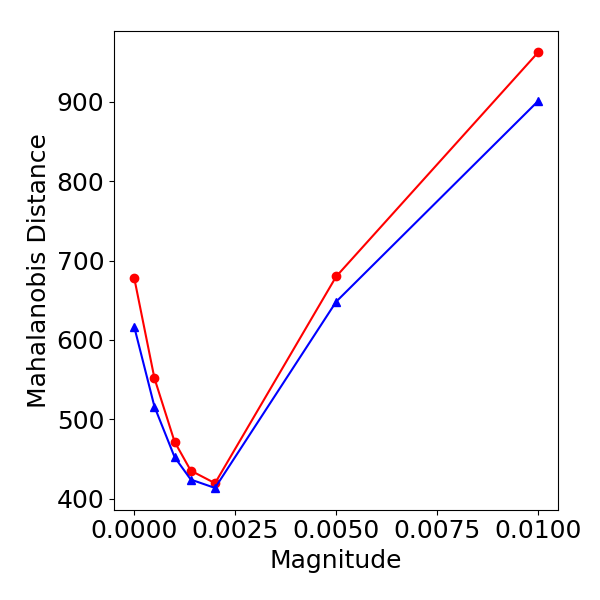}} &
\bmvaHangBox{\includegraphics[height=1.in]{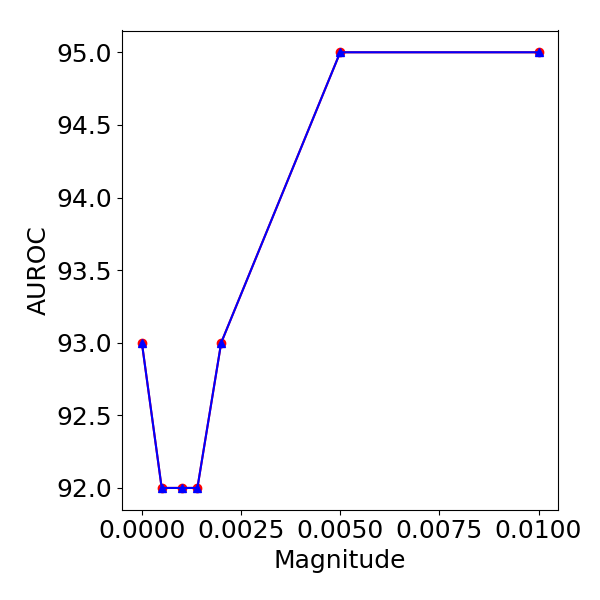}} \\
(a) L1 Norm & (b) CIFAR-10 & (c) SVHN & (d) AUROC
\end{tabular}
\caption{Model: ResNet In: CIFAR-10 Out: SVHN}
\label{fig:resnet_cifar10_svhn}
\vskip 1em
\center
\begin{tabular}{cccc}
\bmvaHangBox{\includegraphics[height=1.in]{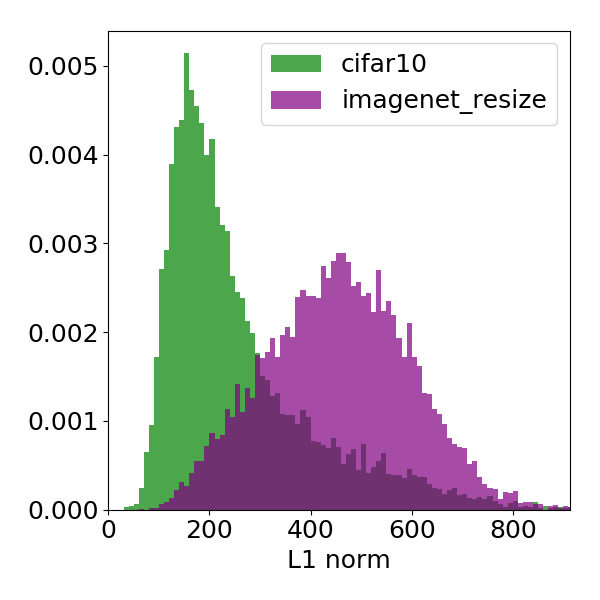}} &
\bmvaHangBox{\includegraphics[height=1.in]{figs/input_preprocessing/resnet_cifar10_in}} &
\bmvaHangBox{\includegraphics[height=1.in]{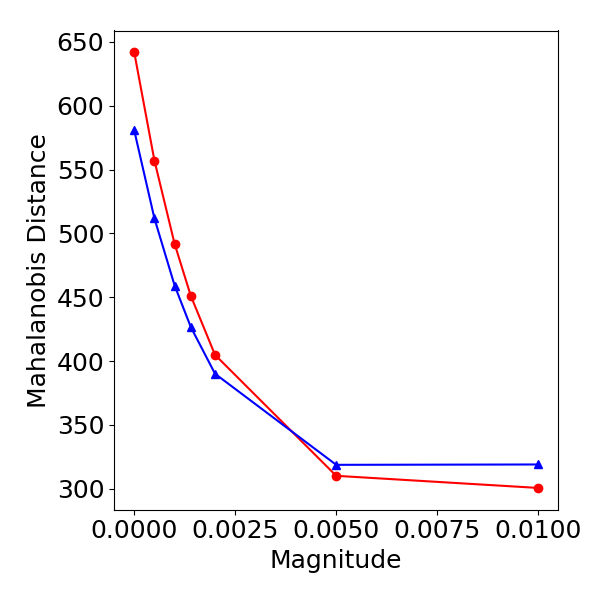}} &
\bmvaHangBox{\includegraphics[height=1.in]{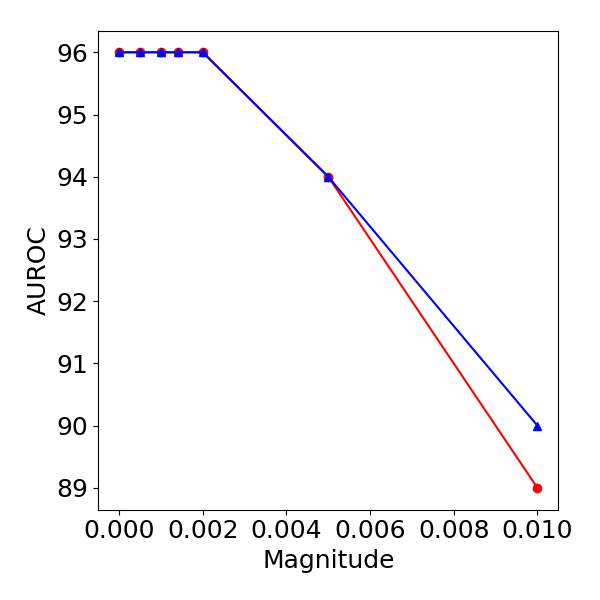}} \\
(a) L1 Norm & (b) CIFAR-10 & (c) ImageNet & (d) AUROC
\end{tabular}
\caption{Model: ResNet In: CIFAR-10 Out: ImageNet}
\label{fig:resnet_cifar10_imagenet_resize}
\vskip 1em
\center
\begin{tabular}{cccc}
\bmvaHangBox{\includegraphics[height=1.in]{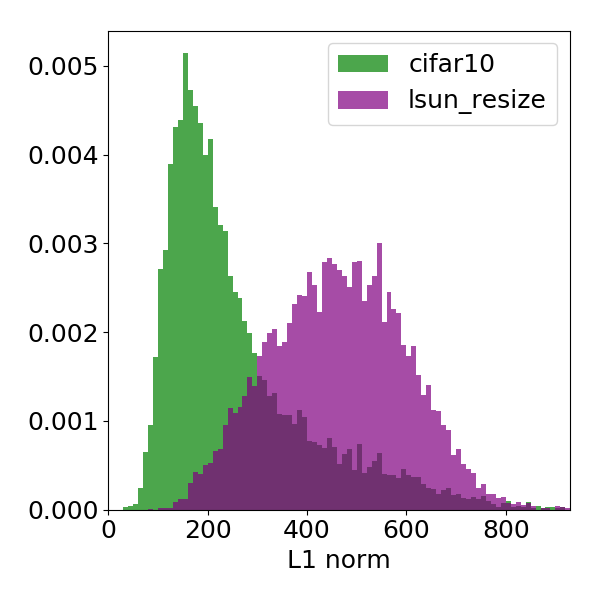}} &
\bmvaHangBox{\includegraphics[height=1.in]{figs/input_preprocessing/resnet_cifar10_in}} &
\bmvaHangBox{\includegraphics[height=1.in]{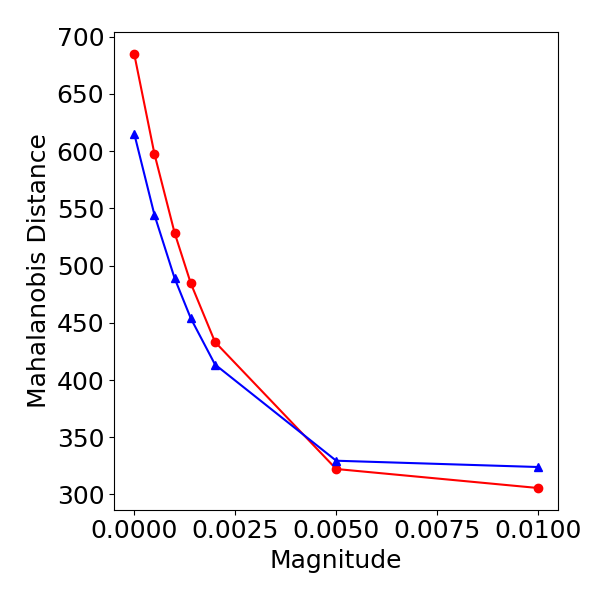}} &
\bmvaHangBox{\includegraphics[height=1.in]{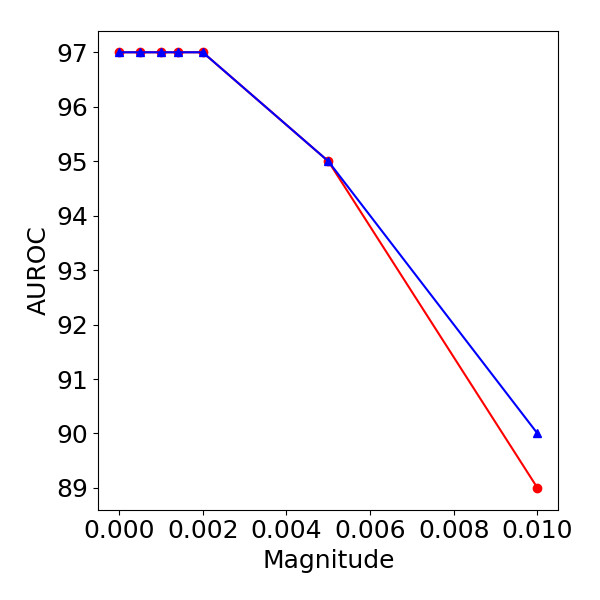}} \\
(a) L1 Norm & (b) CIFAR-10 & (c) LSUN & (d) AUROC
\end{tabular}
\caption{Model: ResNet In: CIFAR-10 Out: LSUN}
\label{fig:resnet_cifar10_lsun_resize}
\vskip 1em
\center
\begin{tabular}{cccc}
\bmvaHangBox{\includegraphics[height=1.in]{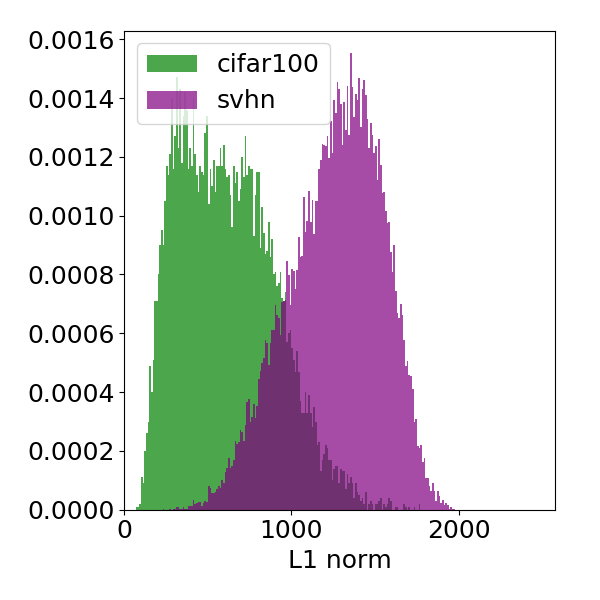}} &
\bmvaHangBox{\includegraphics[height=1.in]{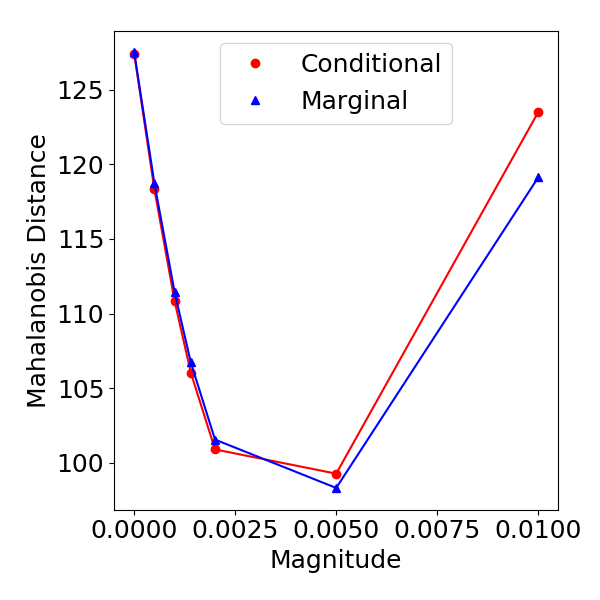}} &
\bmvaHangBox{\includegraphics[height=1.in]{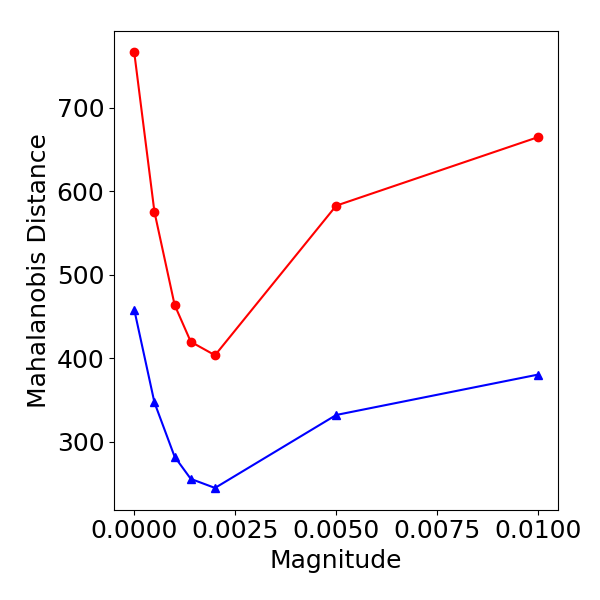}} &
\bmvaHangBox{\includegraphics[height=1.in]{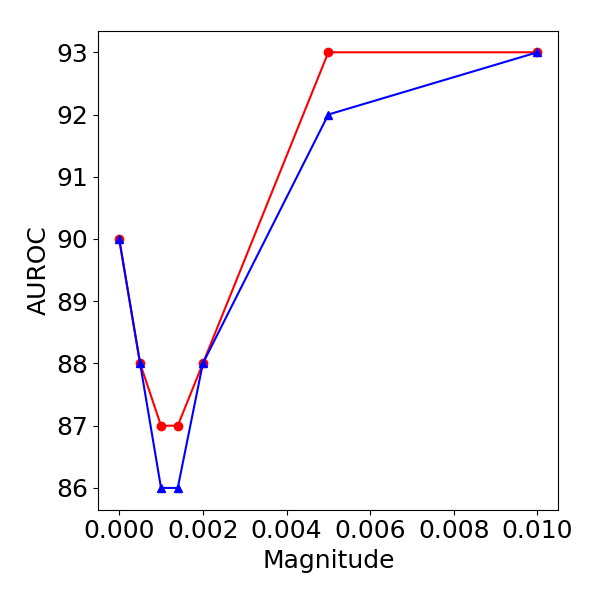}} \\
(a) L1 Norm & (b) CIFAR-100 & (c) SVHN & (d) AUROC
\end{tabular}
\caption{Model: ResNet In: CIFAR-100 Out: SVHN}
\label{fig:resnet_cifar100_svhn}
\vskip 1em
\center
\begin{tabular}{cccc}
\bmvaHangBox{\includegraphics[height=1.in]{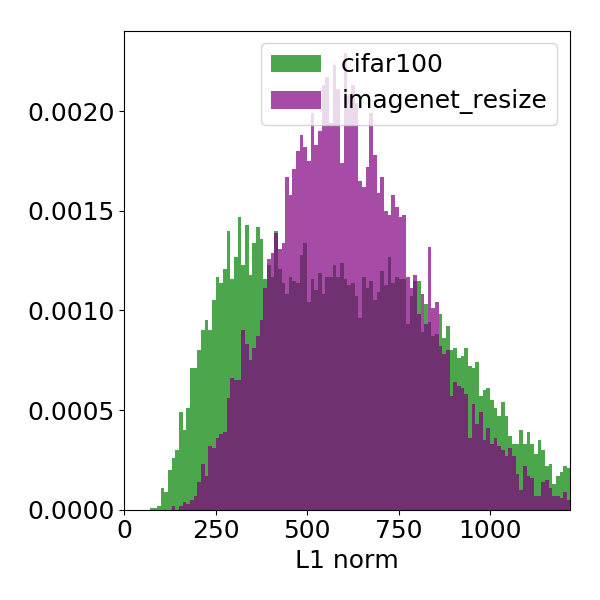}} &
\bmvaHangBox{\includegraphics[height=1.in]{figs/input_preprocessing/resnet_cifar100_in}} &
\bmvaHangBox{\includegraphics[height=1.in]{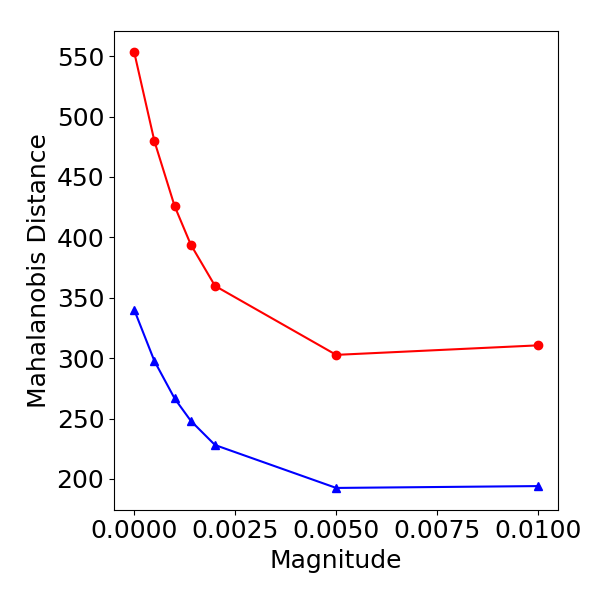}} &
\bmvaHangBox{\includegraphics[height=1.in]{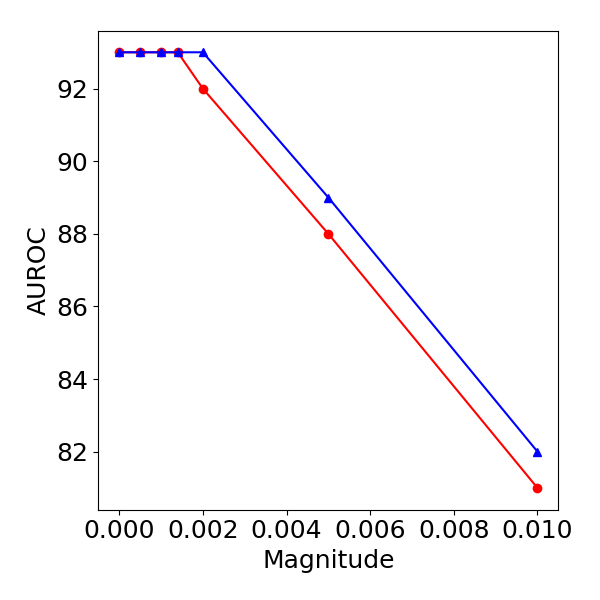}} \\
(a) L1 Norm & (b) CIFAR-100 & (c) ImageNet & (d) AUROC
\end{tabular}
\caption{Model: ResNet In: CIFAR-100 Out: ImageNet}
\label{fig:resnet_cifar100_imagenet_resize}
\vskip 1em
\end{figure}
\begin{figure}[tb]
\center
\begin{tabular}{cccc}
\bmvaHangBox{\includegraphics[height=1.in]{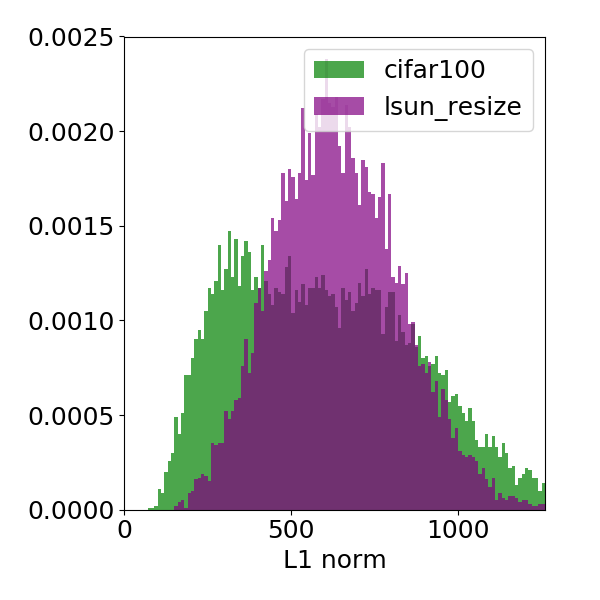}} &
\bmvaHangBox{\includegraphics[height=1.in]{figs/input_preprocessing/resnet_cifar100_in}} &
\bmvaHangBox{\includegraphics[height=1.in]{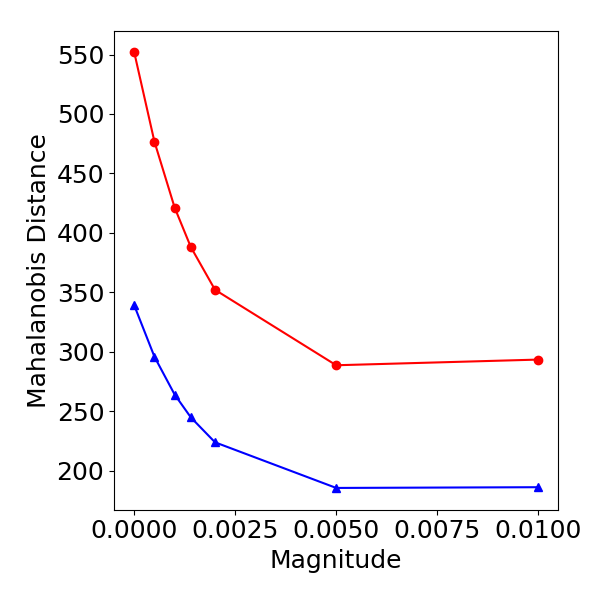}} &
\bmvaHangBox{\includegraphics[height=1.in]{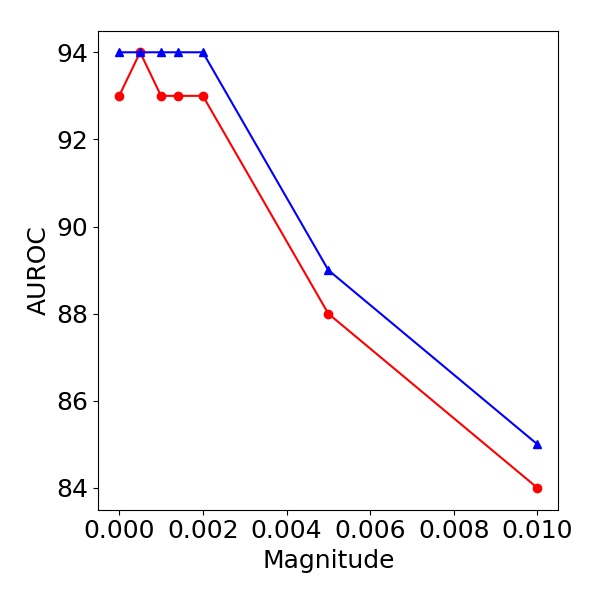}} \\
(a) L1 Norm & (b) CIFAR-100 & (c) LSUN & (d) AUROC
\end{tabular}
\caption{Model: ResNet In: CIFAR-100 Out: LSUN}
\label{fig:resnet_cifar100_lsun_resize}
\vskip 1em
\center
\begin{tabular}{cccc}
\bmvaHangBox{\includegraphics[height=1.in]{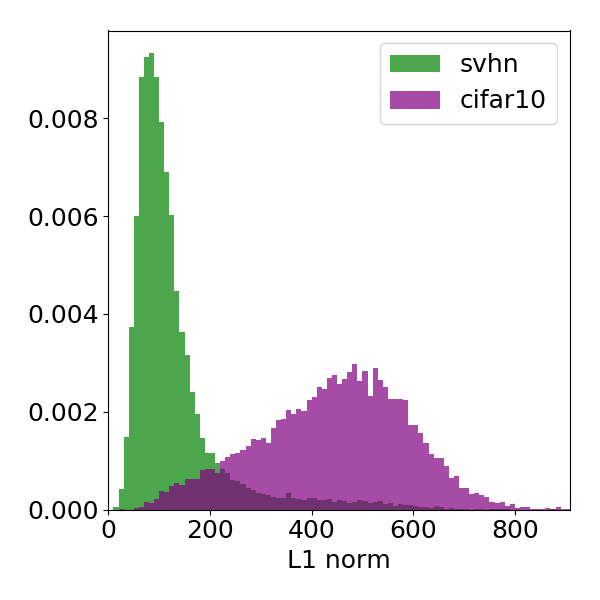}} &
\bmvaHangBox{\includegraphics[height=1.in]{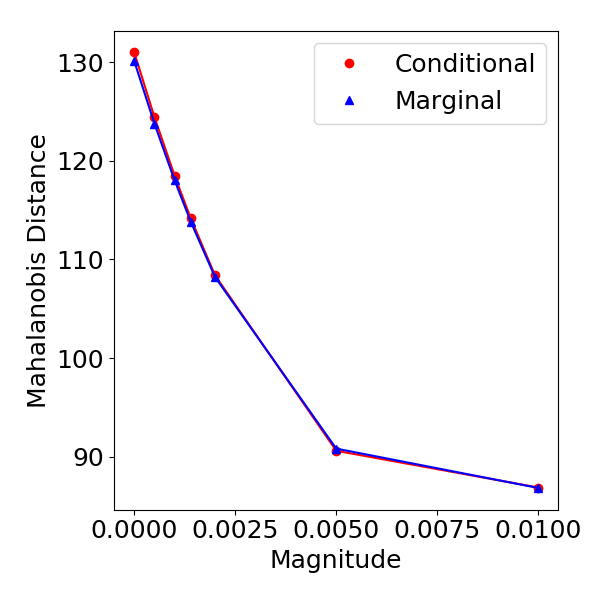}} &
\bmvaHangBox{\includegraphics[height=1.in]{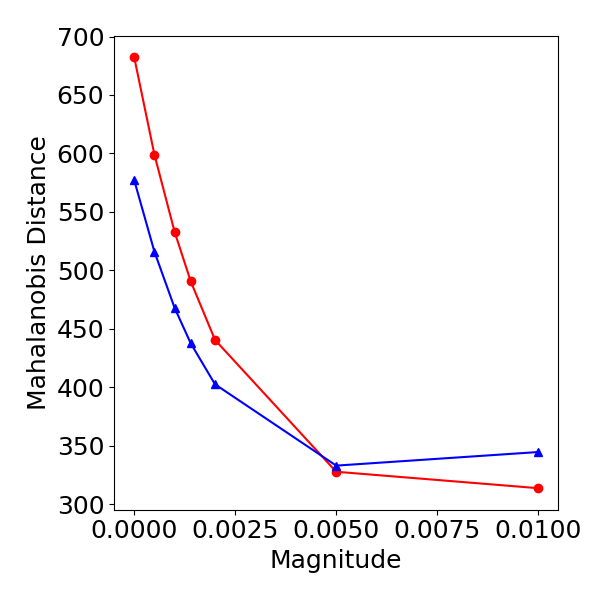}} &
\bmvaHangBox{\includegraphics[height=1.in]{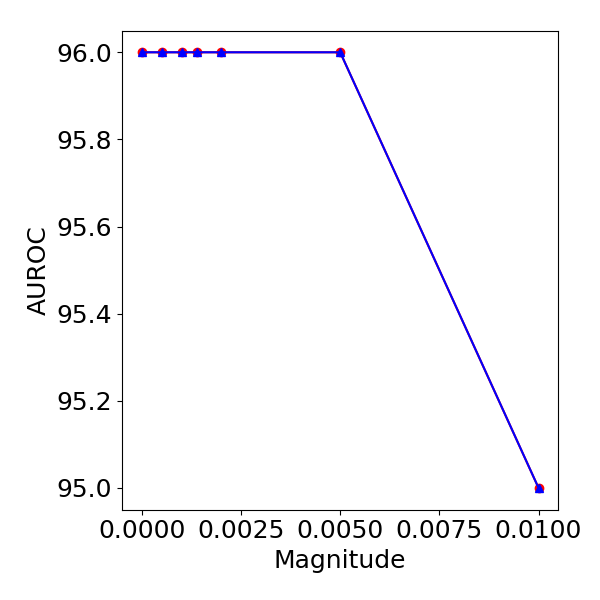}} \\
(a) L1 Norm & (b) SVHN & (c) CIFAR-10 & (d) AUROC
\end{tabular}
\caption{Model: ResNet In: SVHN Out: CIFAR-10}
\label{fig:resnet_svhn_cifar10}
\vskip 1em
\center
\begin{tabular}{cccc}
\bmvaHangBox{\includegraphics[height=1.in]{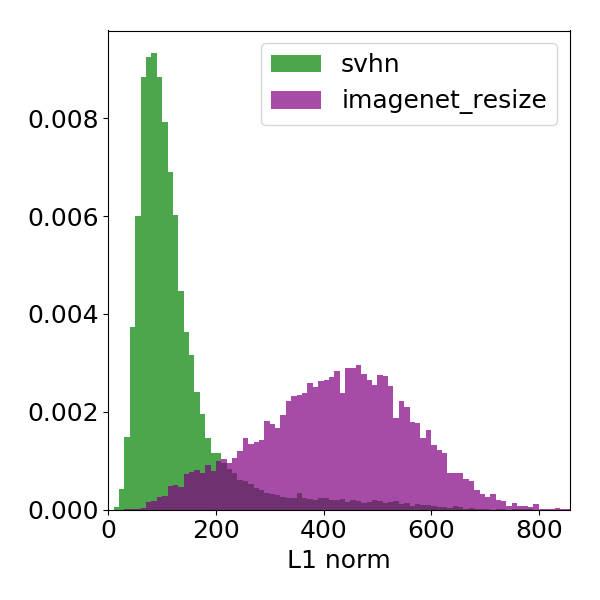}} &
\bmvaHangBox{\includegraphics[height=1.in]{figs/input_preprocessing/resnet_svhn_in}} &
\bmvaHangBox{\includegraphics[height=1.in]{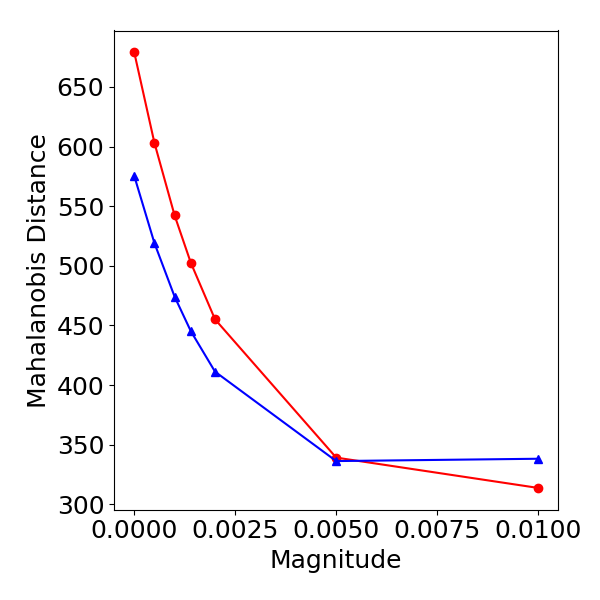}} &
\bmvaHangBox{\includegraphics[height=1.in]{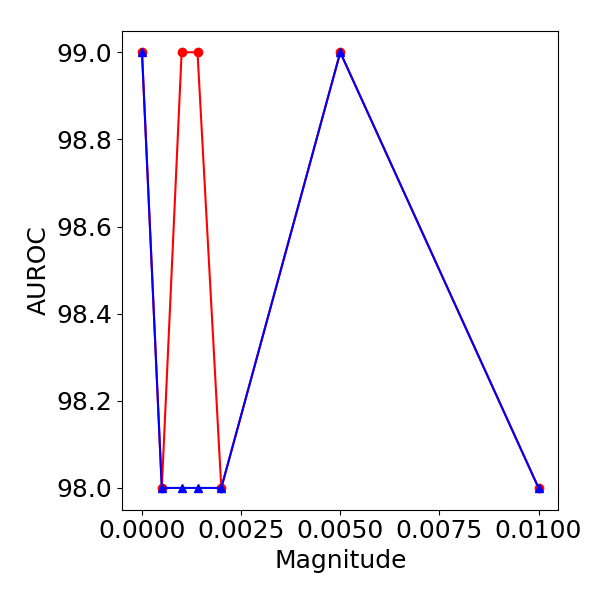}} \\
(a) L1 Norm & (b) SVHN & (c) ImageNet & (d) AUROC
\end{tabular}
\caption{Model: ResNet In: SVHN Out: ImageNet}
\label{fig:resnet_svhn_imagenet_resize}
\vskip 1em
\center
\begin{tabular}{cccc}
\bmvaHangBox{\includegraphics[height=1.in]{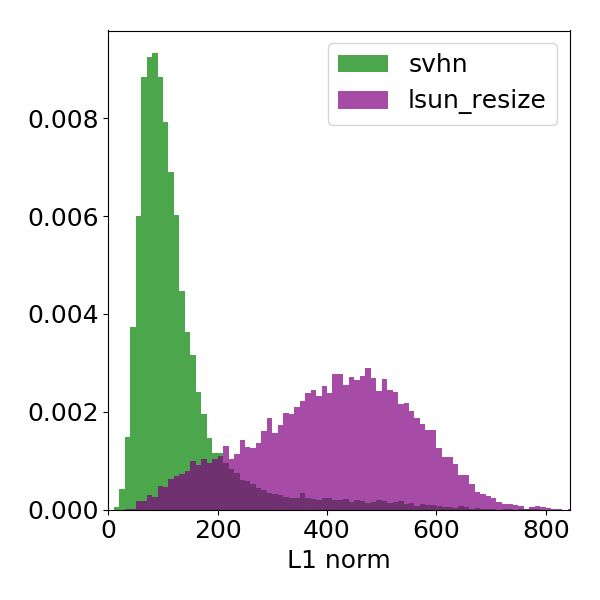}} &
\bmvaHangBox{\includegraphics[height=1.in]{figs/input_preprocessing/resnet_svhn_in}} &
\bmvaHangBox{\includegraphics[height=1.in]{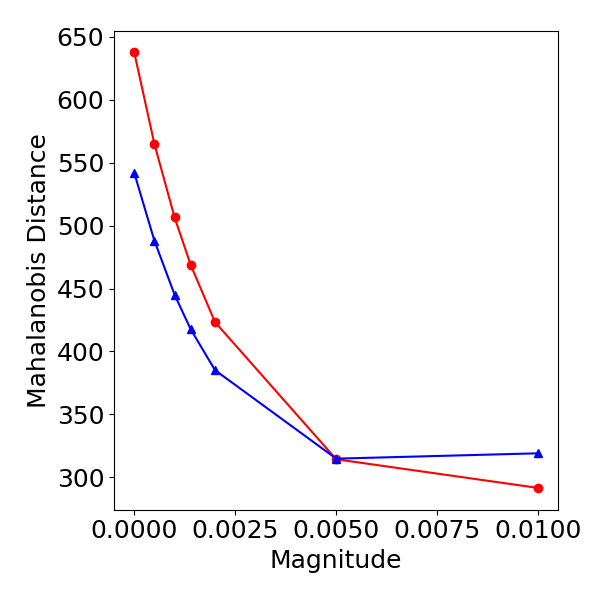}} &
\bmvaHangBox{\includegraphics[height=1.in]{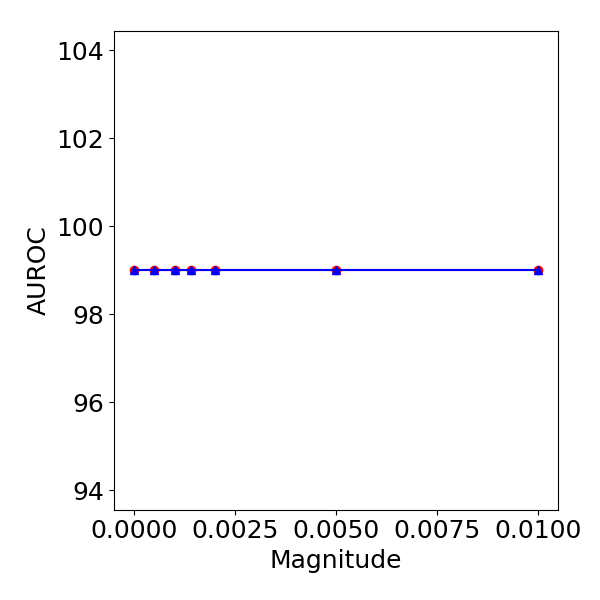}} \\
(a) L1 Norm & (b) SVHN & (c) LSUN & (d) AUROC
\end{tabular}
\caption{Model: ResNet In: SVHN Out: LSUN}
\label{fig:resnet_svhn_lsun_resize}
\vskip 1em
\center
\begin{tabular}{cccc}
\bmvaHangBox{\includegraphics[height=1.in]{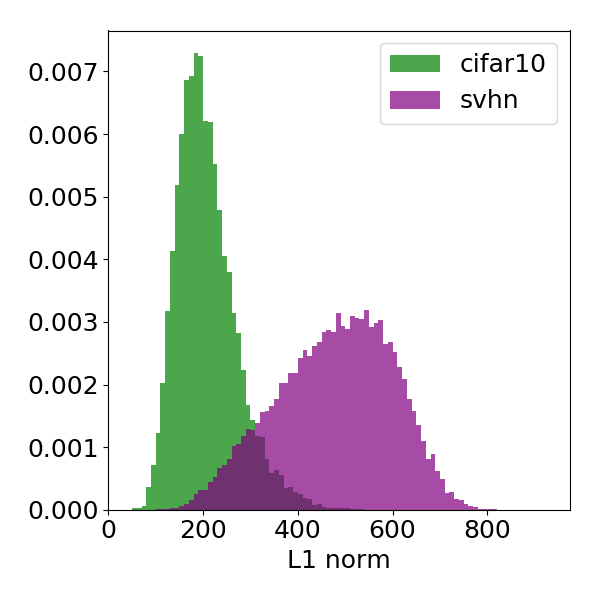}} &
\bmvaHangBox{\includegraphics[height=1.in]{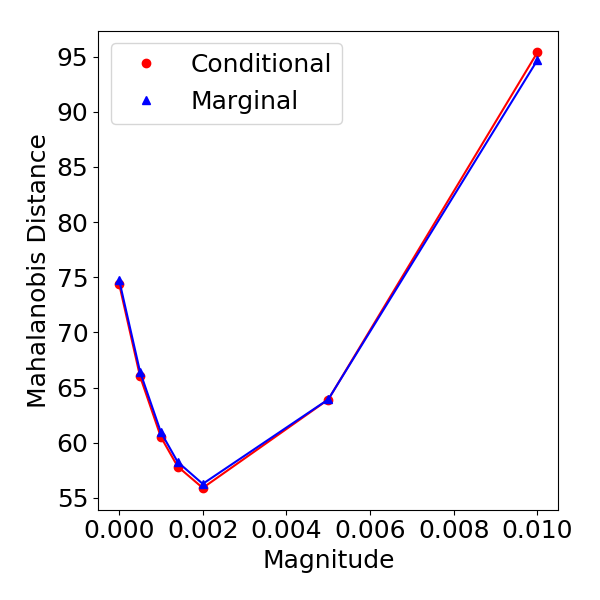}} &
\bmvaHangBox{\includegraphics[height=1.in]{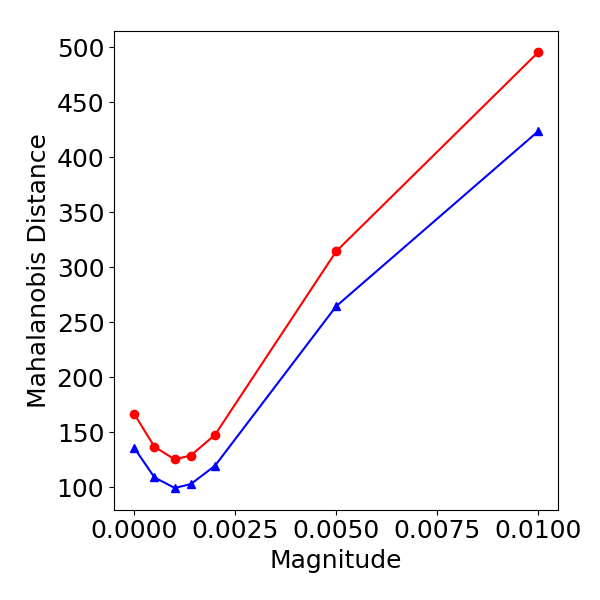}} &
\bmvaHangBox{\includegraphics[height=1.in]{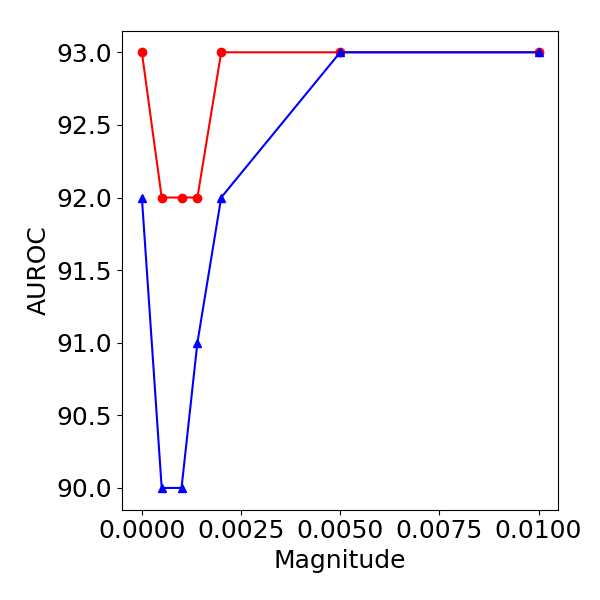}} \\
(a) L1 Norm & (b) CIFAR-10 & (c) SVHN & (d) AUROC
\end{tabular}
\caption{Model: DenseNet In: CIFAR-10 Out: SVHN}
\label{fig:densenet_cifar10_svhn}
\vskip 1em
\end{figure}
\begin{figure}[tb]
\center
\begin{tabular}{cccc}
\bmvaHangBox{\includegraphics[height=1.in]{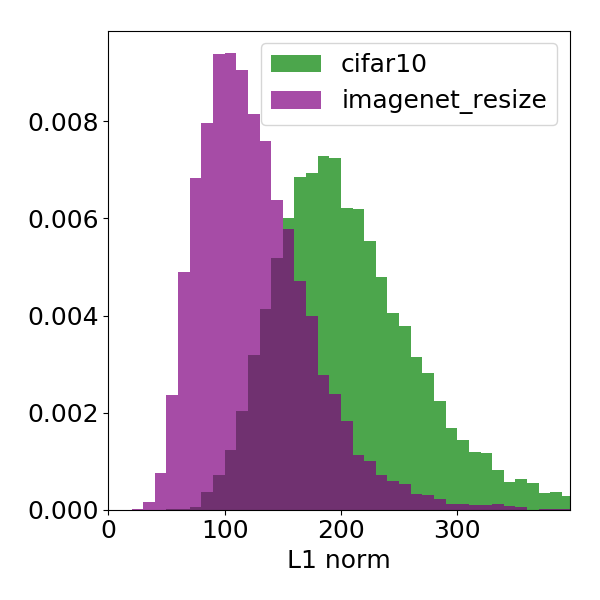}} &
\bmvaHangBox{\includegraphics[height=1.in]{figs/input_preprocessing/densenet_cifar10_in}} &
\bmvaHangBox{\includegraphics[height=1.in]{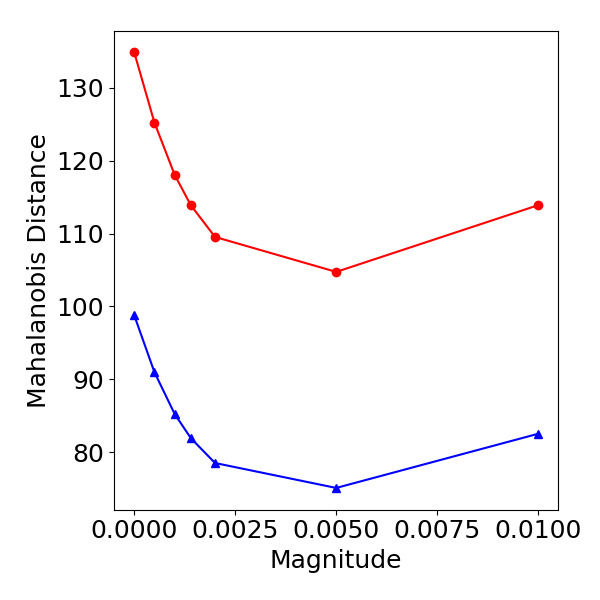}} &
\bmvaHangBox{\includegraphics[height=1.in]{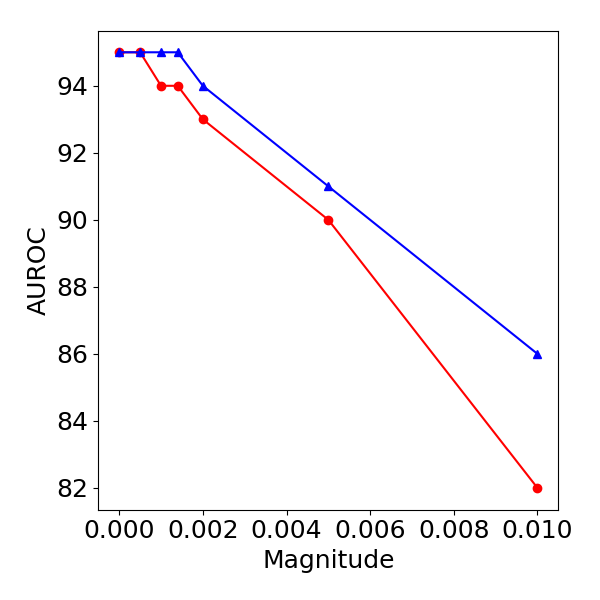}} \\
(a) L1 Norm & (b) CIFAR-10 & (c) ImageNet & (d) AUROC
\end{tabular}
\caption{Model: DenseNet In: CIFAR-10 Out: ImageNet}
\label{fig:densenet_cifar10_imagenet_resize}
\vskip 1em
\center
\begin{tabular}{cccc}
\bmvaHangBox{\includegraphics[height=1.in]{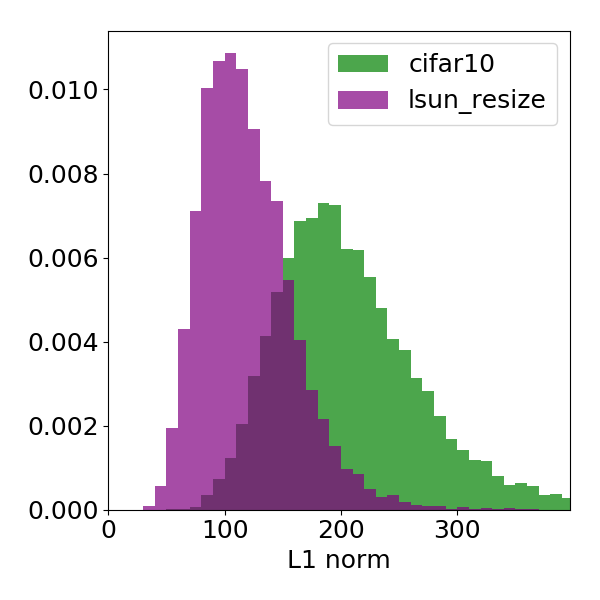}} &
\bmvaHangBox{\includegraphics[height=1.in]{figs/input_preprocessing/densenet_cifar10_in}} &
\bmvaHangBox{\includegraphics[height=1.in]{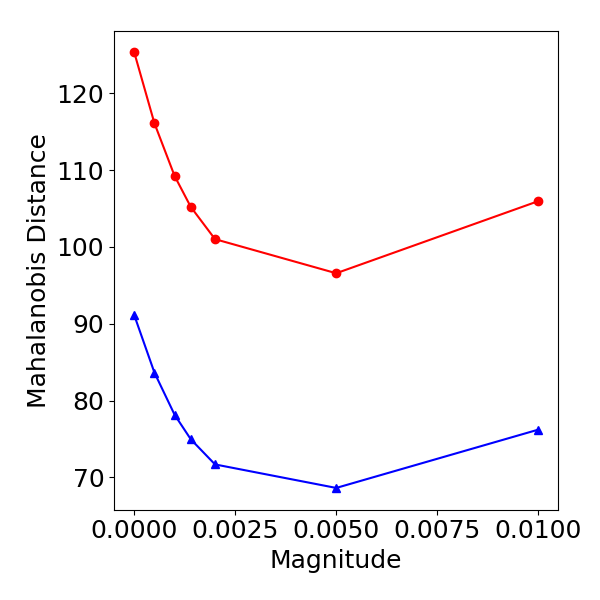}} &
\bmvaHangBox{\includegraphics[height=1.in]{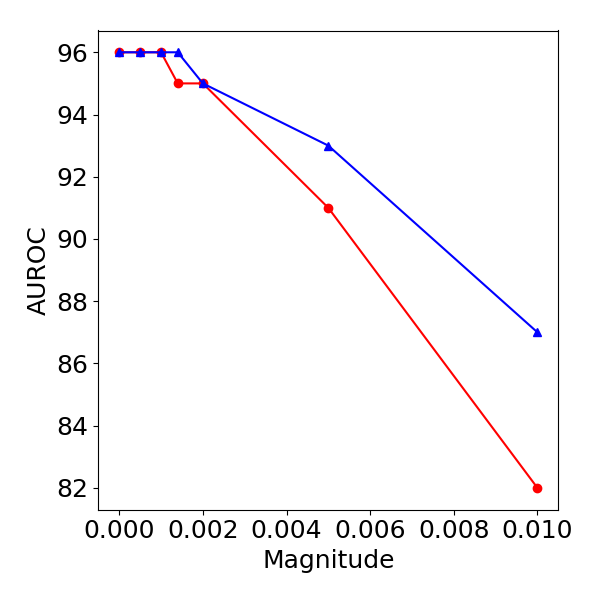}} \\
(a) L1 Norm & (b) CIFAR-10 & (c) LSUN & (d) AUROC
\end{tabular}
\caption{Model: DenseNet In: CIFAR-10 Out: LSUN}
\label{fig:densenet_cifar10_lsun_resize}
\vskip 1em
\center
\begin{tabular}{cccc}
\bmvaHangBox{\includegraphics[height=1.in]{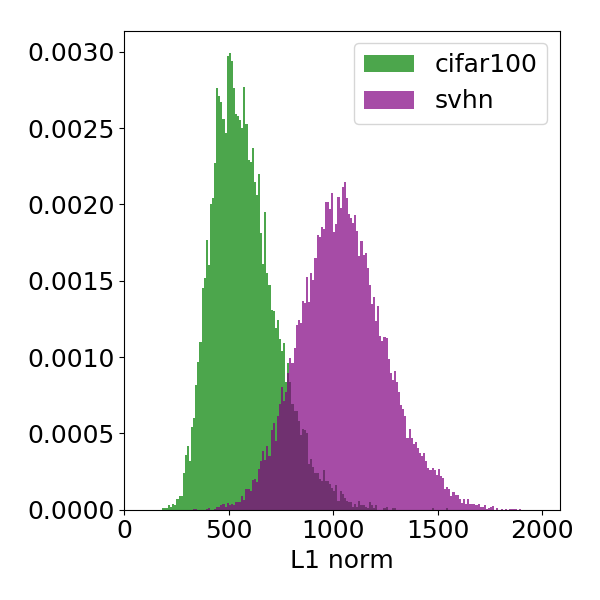}} &
\bmvaHangBox{\includegraphics[height=1.in]{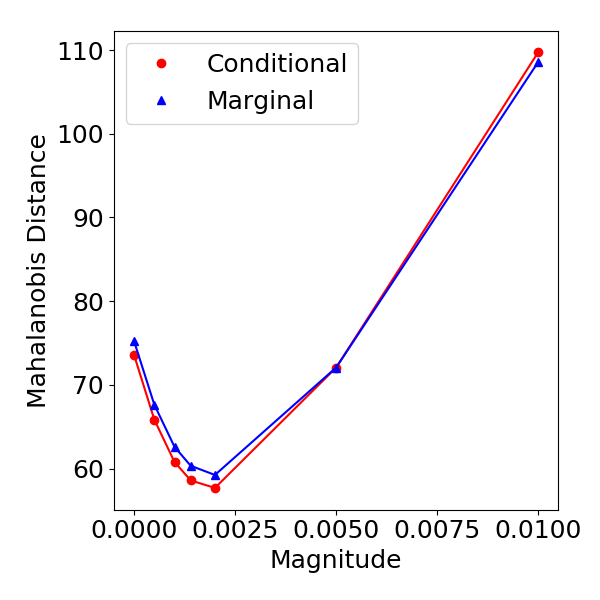}} &
\bmvaHangBox{\includegraphics[height=1.in]{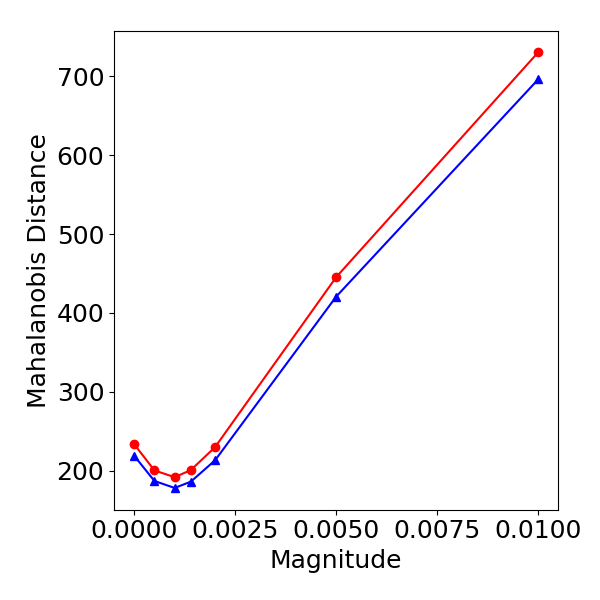}} &
\bmvaHangBox{\includegraphics[height=1.in]{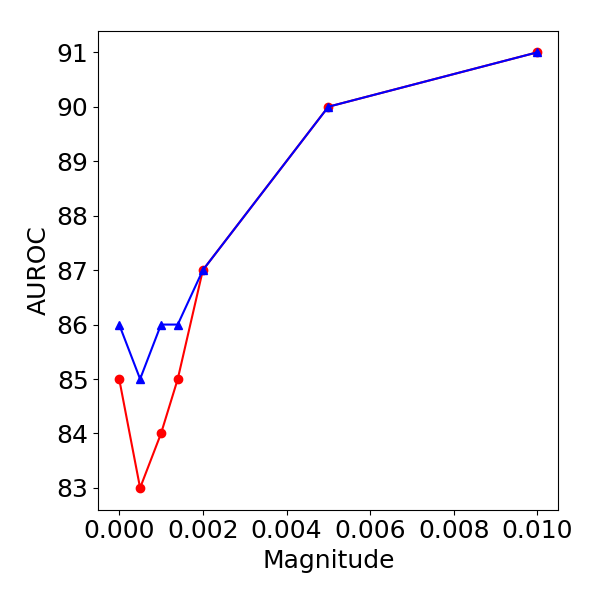}} \\
(a) L1 Norm & (b) CIFAR-100 & (c) SVHN & (d) AUROC
\end{tabular}
\caption{Model: DenseNet In: CIFAR-100 Out: SVHN}
\label{fig:densenet_cifar100_svhn}
\vskip 1em
\center
\begin{tabular}{cccc}
\bmvaHangBox{\includegraphics[height=1.in]{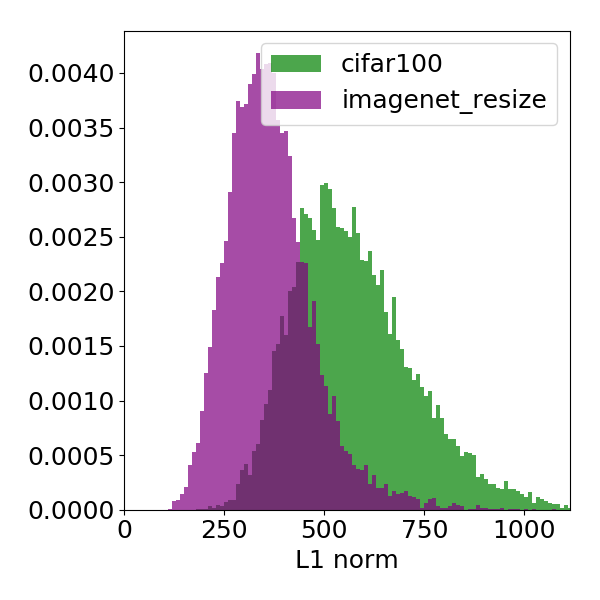}} &
\bmvaHangBox{\includegraphics[height=1.in]{figs/input_preprocessing/densenet_cifar100_in}} &
\bmvaHangBox{\includegraphics[height=1.in]{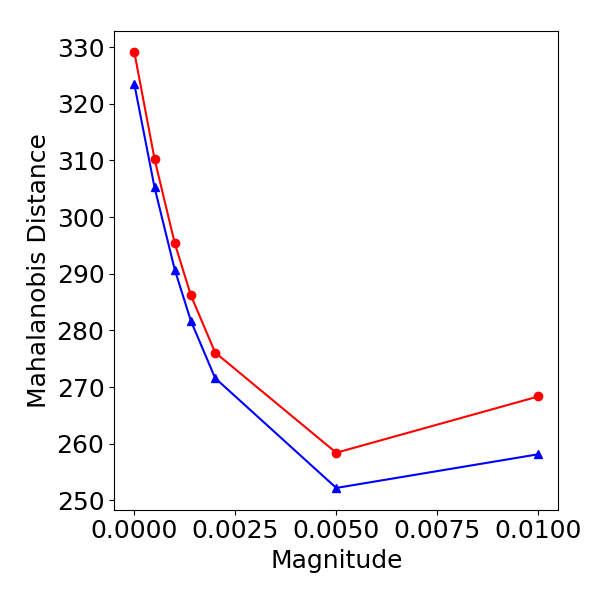}} &
\bmvaHangBox{\includegraphics[height=1.in]{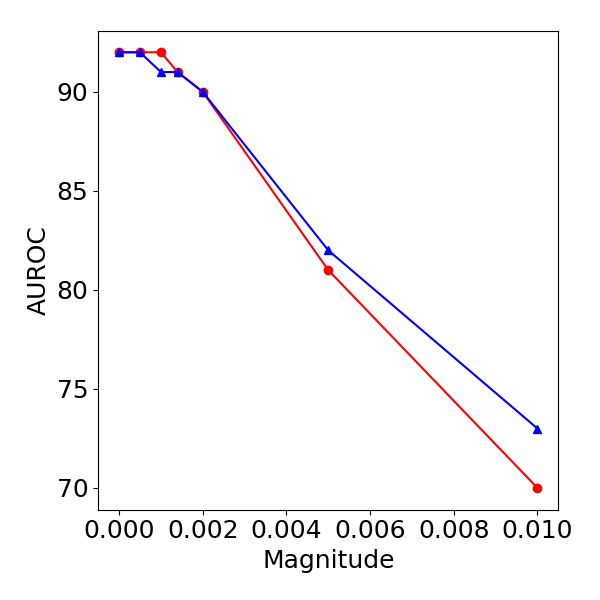}} \\
(a) L1 Norm & (b) CIFAR-100 & (c) ImageNet & (d) AUROC
\end{tabular}
\caption{Model: DenseNet In: CIFAR-100 Out: ImageNet}
\label{fig:densenet_cifar100_imagenet_resize}
\vskip 1em
\center
\begin{tabular}{cccc}
\bmvaHangBox{\includegraphics[height=1.in]{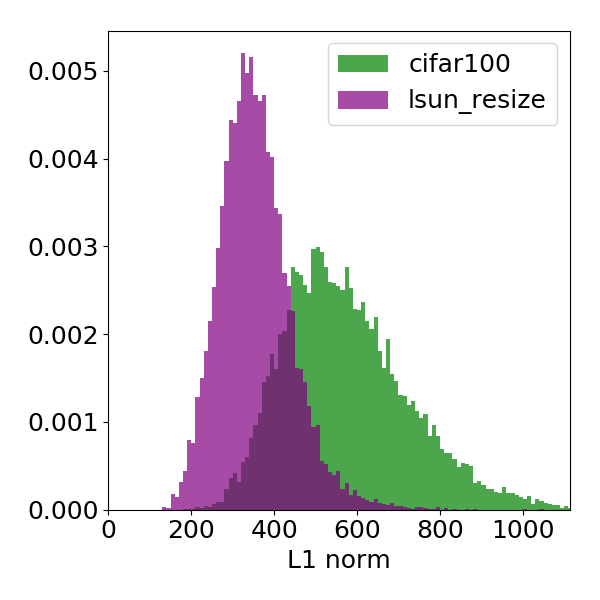}} &
\bmvaHangBox{\includegraphics[height=1.in]{figs/input_preprocessing/densenet_cifar100_in}} &
\bmvaHangBox{\includegraphics[height=1.in]{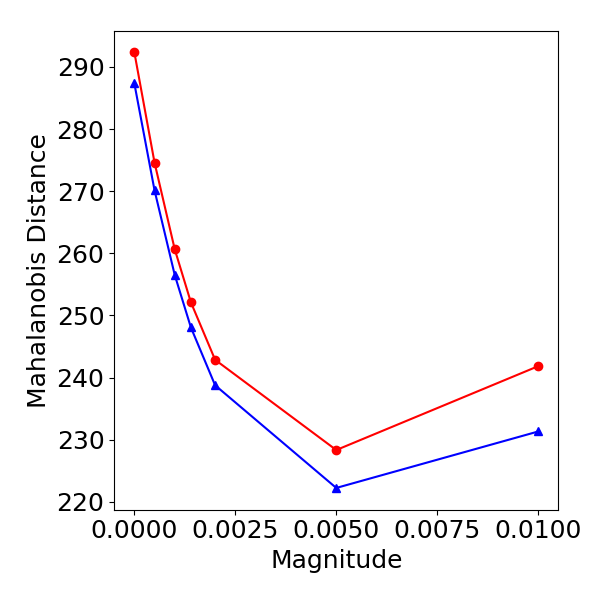}} &
\bmvaHangBox{\includegraphics[height=1.in]{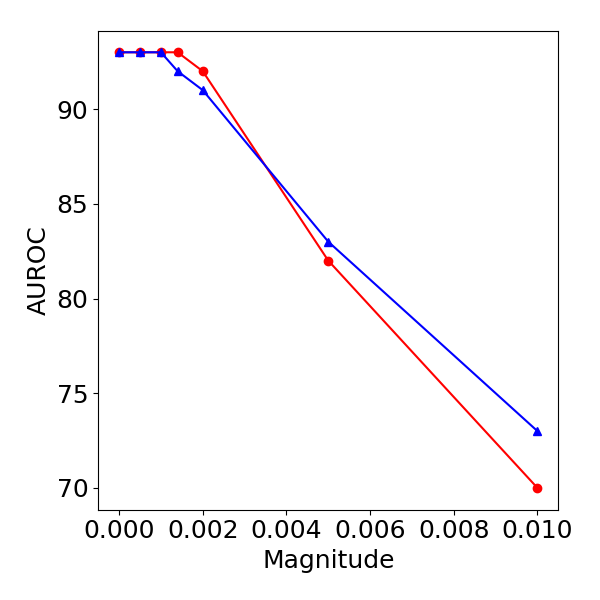}} \\
(a) L1 Norm & (b) CIFAR-100 & (c) LSUN & (d) AUROC
\end{tabular}
\caption{Model: DenseNet In: CIFAR-100 Out: LSUN}
\label{fig:densenet_cifar100_lsun_resize}
\vskip 1em
\end{figure}
\begin{figure}[tb]
\center
\begin{tabular}{cccc}
\bmvaHangBox{\includegraphics[height=1.in]{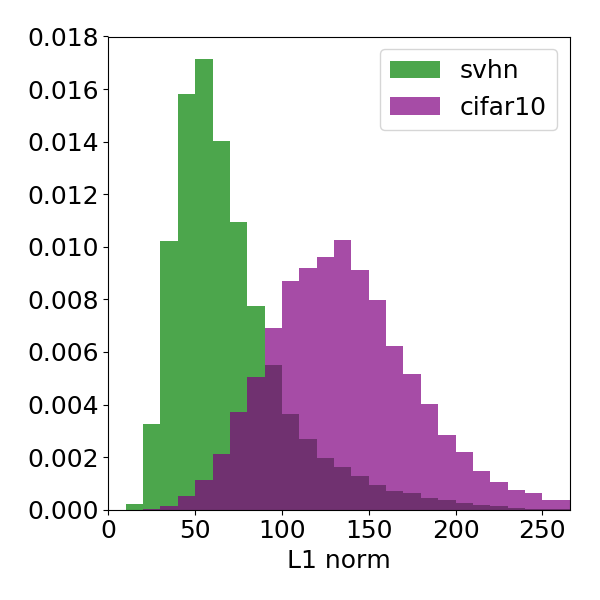}} &
\bmvaHangBox{\includegraphics[height=1.in]{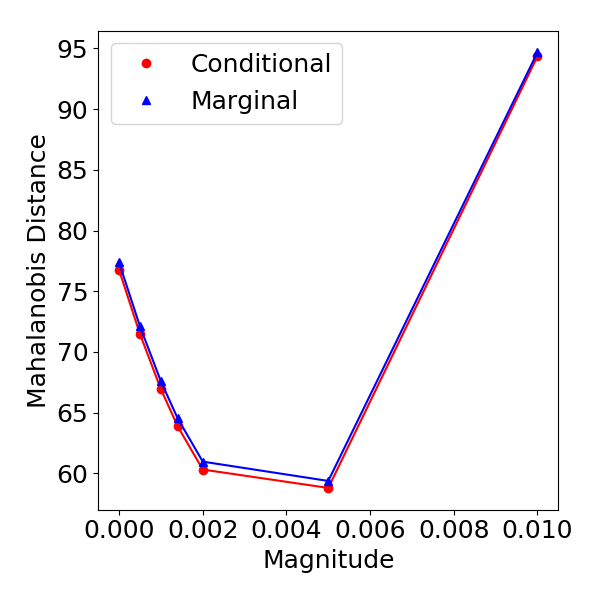}} &
\bmvaHangBox{\includegraphics[height=1.in]{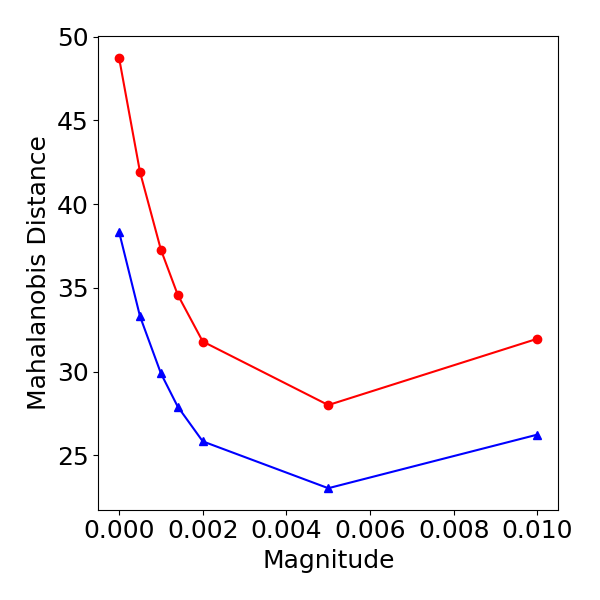}} &
\bmvaHangBox{\includegraphics[height=1.in]{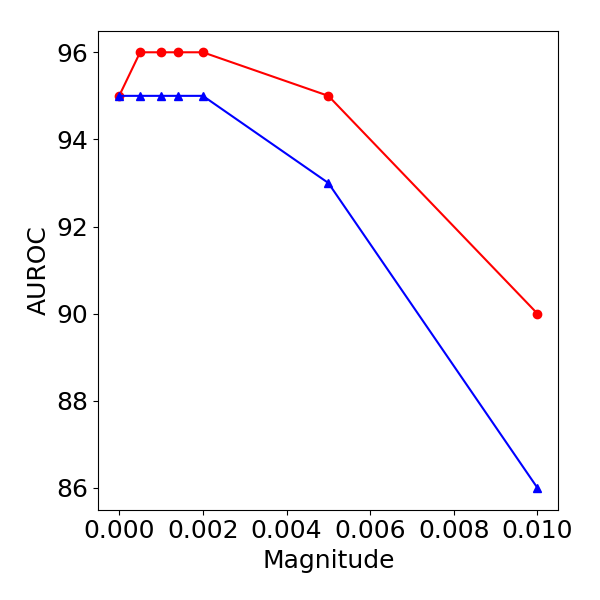}} \\
(a) L1 Norm & (b) SVHN & (c) CIFAR-10 & (d) AUROC
\end{tabular}
\caption{Model: DenseNet In: SVHN Out: CIFAR-10}
\label{fig:densenet_svhn_cifar10}
\vskip 1em
\center
\begin{tabular}{cccc}
\bmvaHangBox{\includegraphics[height=1.in]{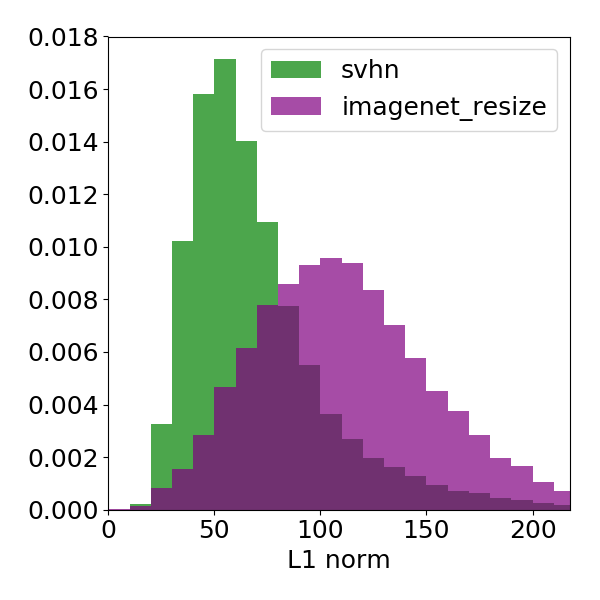}} &
\bmvaHangBox{\includegraphics[height=1.in]{figs/input_preprocessing/densenet_svhn_in}} &
\bmvaHangBox{\includegraphics[height=1.in]{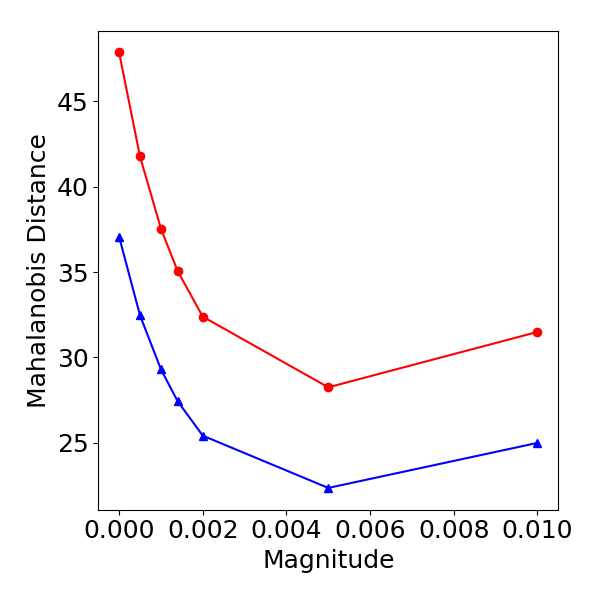}} &
\bmvaHangBox{\includegraphics[height=1.in]{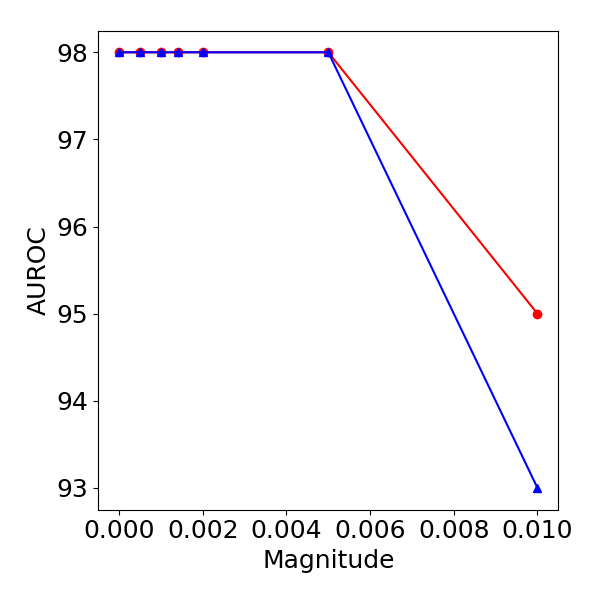}} \\
(a) L1 Norm & (b) SVHN & (c) ImageNet & (d) AUROC
\end{tabular}
\caption{Model: DenseNet In: SVHN Out: ImageNet}
\label{fig:densenet_svhn_imagenet_resize}
\vskip 1em
\center
\begin{tabular}{cccc}
\bmvaHangBox{\includegraphics[height=1.in]{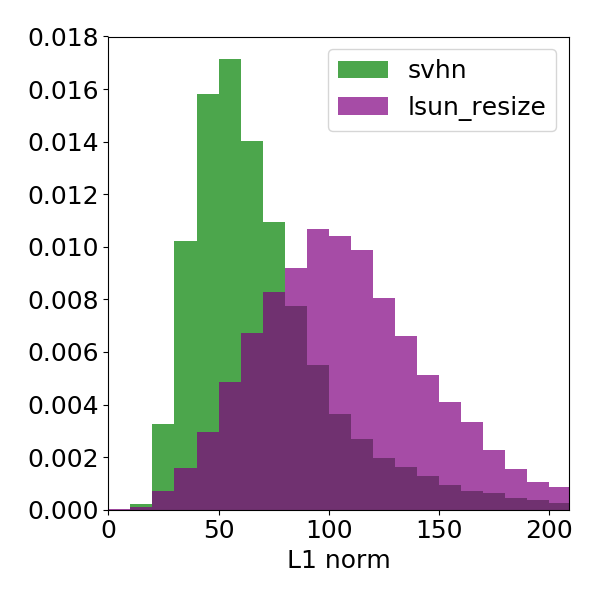}} &
\bmvaHangBox{\includegraphics[height=1.in]{figs/input_preprocessing/densenet_svhn_in}} &
\bmvaHangBox{\includegraphics[height=1.in]{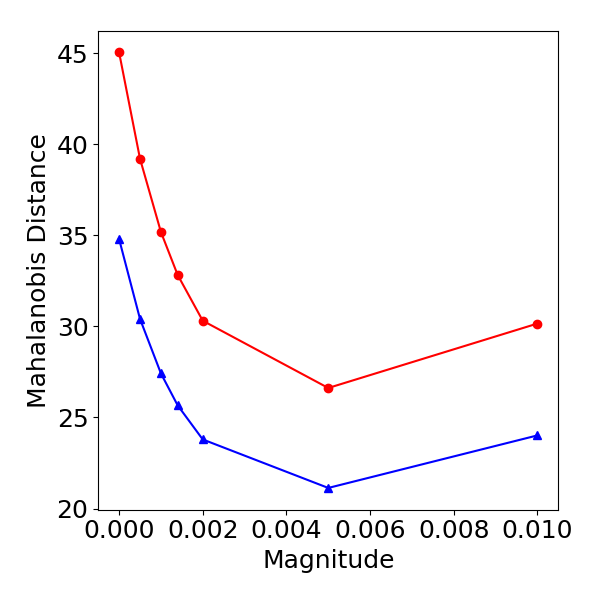}} &
\bmvaHangBox{\includegraphics[height=1.in]{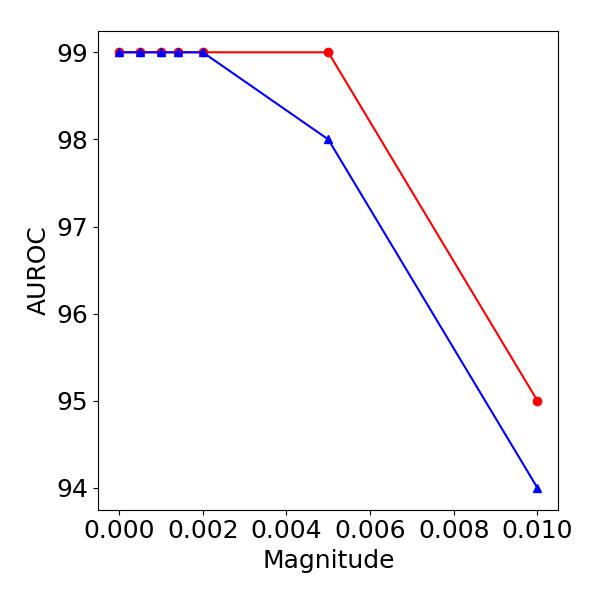}} \\
(a) L1 Norm & (b) SVHN & (c) LSUN & (d) AUROC
\end{tabular}
\caption{Model: DenseNet In: SVHN Out: LSUN}
\label{fig:densenet_svhn_lsun_resize}
\vskip 1em
\end{figure}
 \fi

\end{document}